\documentclass[pdflatex,oneside]{sn-jnl}

\usepackage{graphicx}%
\usepackage{multirow}%
\usepackage{amsmath,amssymb,amsfonts}%
\usepackage{amsthm}%
\usepackage{mathrsfs}%
\usepackage[title]{appendix}%
\usepackage{xcolor}%
\usepackage{textcomp}%
\usepackage{manyfoot}%
\usepackage{booktabs}%
\usepackage{algorithm}%
\usepackage{algorithmicx}%
\usepackage{algpseudocode}%
\usepackage{listings}%
\usepackage{subcaption}
\usepackage{makecell}
\usepackage{placeins}
\usepackage{natbib}

\theoremstyle{thmstyleone}%

\theoremstyle{thmstyletwo}%

\theoremstyle{thmstylethree}%

\begin{document}

\title[Modeling Components and Connections in Cyber-Physical Systems]{Modeling Components and Connections in Cyber-Physical Systems}

\author[1]{\fnm{Kate} \sur{Sanborn}}\email{kate.l.sanborn@vanderbilt.edu}

\author[1]{\fnm{Tanuj} \sur{Kenchannavar}}\email{tanuj.g.kenchannavar@vanderbilt.edu}

\author[1]{\fnm{Vakul} \sur{Nath}}\email{vakul.nath@vanderbilt.edu}

\author[1]{\fnm{Jonathan} \sur{Sprinkle}}\email{jonathan.sprinkle@vanderbilt.edu}

\affil[1]{\orgdiv{Department of Computer Science}, \orgname{Vanderbilt University}, \orgaddress{\street{Vanderbilt Place}, \city{Nashville}, \postcode{37235}, \state{Tennessee}, \country{United States}}}

\abstract
{Text based configuration files for cyber-physical systems show the hierarchy of component modules well but often hide the details of connections and interfaces between modules. A model-based visual approach to these configuration files can better capture this information. The XML structure of Robot Operating System (ROS) launch files can be improved using a modeling approach.

This paper presents ROSLaunchVisual, a model-integrated environment built on WebGME for designing, visualizing, and managing ROS launch files. The tool raises the level of abstraction by allowing developers to create and modify launch files using a graphical interface that represents nodes, publishers, subscribers, and arguments as interconnected components. The tool provides a dynamic system analysis that can then be used in the static development and analysis of new and existing launch files.
ROSLaunchVisual incorporates features such as metamodel-driven validation, automatic import/export of launch files, and visual communication mapping. Plugins further enhance functionality by updating libraries, checking for semantic errors, and managing remaps. By making launch file creation more intuitive and less error-prone, ROSLaunchVisual improves development efficiency and system understanding, especially in collaborative or large-scale robotics projects.}


\maketitle

\section{Introduction}

Cyber-physical systems are frequently designed as modular components with interconnections. These component-based frameworks model physical systems well. Modular designs often have complexity in the integration of components. A text-based interface is commonly used to define these system configurations. Although textual methods show which elements are present in a system and how these elements are composed hierarchically, it can be challenging for users to understand the interconnection of these different components, as these interfaces are hidden in the textual format.

The following paper presents a model-based tool that provides a visual method of specifying system configurations. The tool provides a dynamic probe of system components and interfaces. These pieces are then loaded into the design environment, allowing the exploration and development of new and existing configuration files in a visual, model-based manner which statically displays potential component connections.

ROSLaunchVisual, the tool presented in this paper, deals with systems built in the Robot Operating System (ROS) \cite{quigley2009ros}. Complex robotic systems can be defined using ROS launch files. ROSLaunchVisual allows users to perform a dynamic evaluation of the properties of the system components and interfaces, load this information into the tool, and then statically explore the design parameters of existing or new launch files. This tool helps bridge the gap between the modular nature of systems and the limits of defining these systems using text-based files. 

\subsection{ROS}
ROS is a flexible framework for writing modular software in robotic systems. Software is divided into single-purpose units called nodes. The nodes are collected into groups called packages. ROS nodes communicate with each other via asynchronous communication through publishers and subscribers. They send messages to each other called topics. Topics have a type and a name. If a publisher and subscriber send and receive topics with the same type and name, the two nodes will connect and communicate.

ROS has a variety of applications in the field of cyber-physical systems \cite{bunting2024libpanda,nice2023middleware,10.1145/3459609.3460531}. The modular nature of ROS allows complex systems to be built by reusing components and setting up configurations. Because of this flexibility, ROS works well for a variety of robotics applications.

\subsection{ROS Launch Files}
Starting ROS nodes in a robotic system can be accomplished in several different ways. One can start each node in its own terminal, one node at a time. This method requires remembering the package and executable name for each node that needs to be started. Also, some nodes require command line arguments to be passed in through the terminal. It can be tedious to start each node individually in this way. Also, this technique makes it difficult to reproduce a set-up, since it may be difficult to remember exactly which nodes were started and what configuration options were used.

ROS launch files \cite{Duan2023old} help eliminate these issues by providing an easier method to start a complex system of ROS nodes. A ROS launch file is an XML file that defines which nodes to start and what configuration options should be used. There are several tags available to be used in a ROS launch file. For example, these include nodes (\lstinline|<node>|), parameters (\lstinline|<param>|), and arguments (\lstinline|<arg>|). A remap tag (\lstinline|<remap>|) changes the name of a publisher or subscriber so that publishers and subscribers with different names can communicate without needing to change the original code. A group tag (\lstinline|<group>|) allows grouping of nodes and other tags so that the scope of configurations is limited to a smaller group. An include tag (\lstinline|<include>|) allows the launch file to start another launch file and everything contained inside of the included file.

Writing a ROS launch file can present several challenges. It can be difficult to type the file and remember the syntax. In \cite{20225213303132}, an analysis of the online Q\&A community revealed that one of the two main problems in writing launch files is syntax errors. It can be difficult to remember all the valid tags, allowed attributes, and nesting rules of the XML file. Another challenge with the XML launch file is that it is difficult to understand how the nodes communicate with each other. Outside of remap tags, the launch file does not have any information about publishers and subscribers in the nodes. If someone is just reading the file, especially if he or she was not the author of the file, it can be tricky to understand and visualize how the nodes are connected in a communication network.

The difficulties in creating a launch file in a text format motivate the project presented in this work. ROSLaunchVisual, a visual solution that demonstrates the communication of nodes, aims to speed up and simplify launch file development by reducing syntax errors and visually displaying how the nodes are connected.

\subsection{Contributions}
This paper is an extension of a previous work \cite{10.1145/3722573.3727832}. The contribution of this project is a tool that allows model-based editing of ROS 1 launch files. ROSLaunchVisual allows importing, editing, and exporting of launch files in a visual format. The tool provides a static analysis of the launch file itself while using information from a dynamic analysis of ROS code to provide information about publishers, subscribers, and topics present in ROS nodes. This journal extension also provides a much deeper analysis of the tool, presenting test results and discussing limitations further.

Section \ref{section:background} provides background on related ROS debugging tools and background on WebGME, the modeling tool that serves as a base for ROSLaunchVisual. Section \ref{section:methods} discusses how the tool works and the different features of the tool that can be used. Section \ref{section:casestudy} gives a complete example of using the tool from start to finish. In Section \ref{section:evaluation}, different features of the tool are evaluated with real-world inputs. The limitations of ROSLaunchVisual are discussed in Section \ref{section:limitations}. Related work is evaluated in Section \ref{section:relatedwork}. Section \ref{section:conclusion} concludes the paper.

\section{Background} \label{section:background}
The following section discusses existing tools that assist with debugging ROS code and WebGME, the base of ROSLaunchVisual. Section \ref{section:existing} describes the tools that exist as part of ROS to assist with understanding the complexity, composition, and communication of a complicated system. Section \ref{section:webgme} discusses WebGME and how it can be useful in modeling.

\subsection{Existing ROS Visualization and Debugging Tools} \label{section:existing}
\begin{figure}
    \centering
    \includegraphics[width=\linewidth]{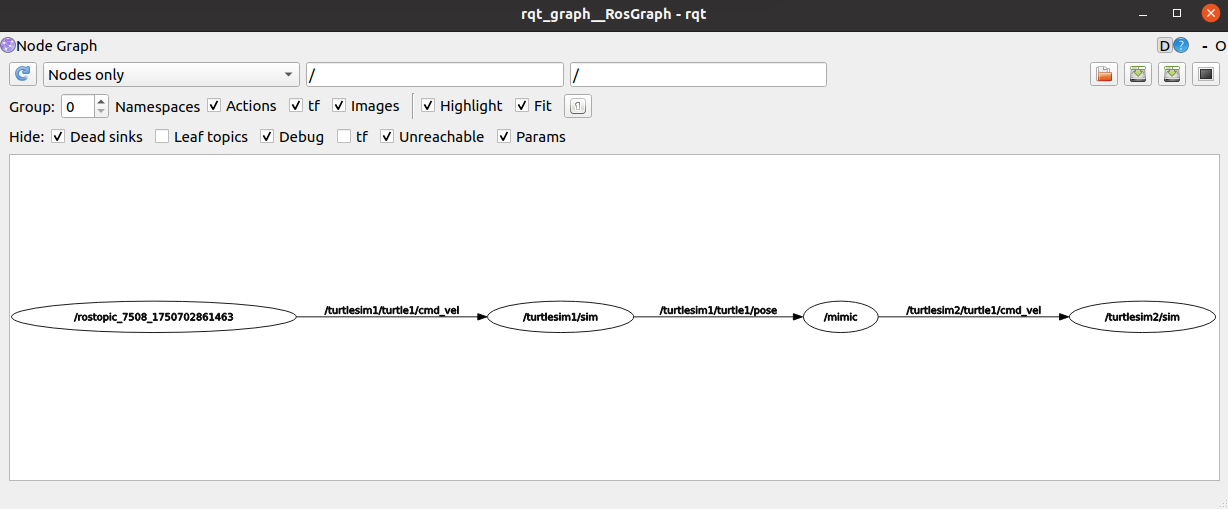}
    \caption{Example RQt display}
    \label{fig:rqt}
\end{figure}

There are several existing ROS tools that can be used to help visualize and debug ROS systems. An example tool is RQt (ROS Qt-based GUI) \cite{Patkar_Mandhalkar_Chavan_Songire_Kothawade_2023}. RQt provides console logging of different messages that a running ROS system produces, such as information messages, warnings, and debug statements. RQt also has graph functionality. The graph shows all  the running nodes. It also shows the topics to which each node publishes or subscribes. An example of an RQt graph is shown in Figure \ref{fig:rqt}. This figure is recreated from the roslaunch and RQt tutorial found in \cite{ros_rqtconsole_tutorial}. It is important to note that this visualization works only for running code. It cannot generate the visual from the static code in a node or launch file. This makes RQt useful for testing applications but less useful in the design phase when building a complicated robotic system.

Another useful tool for debugging ROS systems is roswtf \cite{Duan2023}. This tool can be used to analyze a ROS package or a ROS launch file. It performs some static checks and examines the running system for problems. Static check errors can include incorrect configurations, environment variable problems, and more. With online error analysis, roswtf looks for problems such as unresponsive nodes or missing connections \cite{roswtf}. These errors are presented in words rather than in an image or a diagram.

Both of these existing tools provide useful assistance in testing and debugging code, including understanding how nodes communicate and where communication may fail. However, using these tools to analyze communication requires that the nodes be running. These tools are helpful in testing a finished ROS system, but less useful in the design process. For instance, writing a launch file requires an understanding of the publishers and subscribers on different nodes and how they are connected with each other. These tools can be helpful while writing the launch file to understand the communication of the nodes. However, each time the developer wishes to debug this communication, he or she would have to launch the launch file. This can be tedious and difficult, especially when the launch file is intended to be run on complicated hardware that may be difficult to access for frequent testing. The need for communication visualization during the design phase motivated the creation of the tool presented in this paper.

\subsection{WebGME} \label{section:webgme}

WebGME \cite{kecskes2017bridging} is a meta-programmable modeling environment designed for creating domain-specific modeling languages (DSMLs). It supports collaborative, browser-based model editing with version control and plugin extensibility. In our approach, WebGME is used to construct and validate models of ROS launch files. The metamodel defines allowable components such as nodes, publishers, subscribers, arguments, and remaps, along with their relationships. This enables hierarchical modeling and enforces structural constraints to avoid invalid configurations. Users can visually compose systems using predefined libraries, while plugins handle tasks such as importing XML launch files, updating node libraries, and generating valid ROS output. WebGME’s visual and rule-driven interface enables both novice and expert users to rapidly create and understand ROS system configurations.

\section{Methods} \label{section:methods}

This section describes how ROSLaunchVisual uses metamodeling \cite{sprinkle20073, gray2007domain} and WebGME to build a launch file. Metamodeling is the process of using a metamodel to define the rules, structure, and semantics of a model. WebGME allows developers to design a metamodel and then build models that are instances of the metamodel. Plugins can then be created to assist the design process. The following section describes the metamodel for ROSLaunchVisual and the process of using WebGME and plugins to design and export a functional launch file.

\subsection{Metamodel}
\begin{figure}
    \centering
    \includegraphics[width=\linewidth]{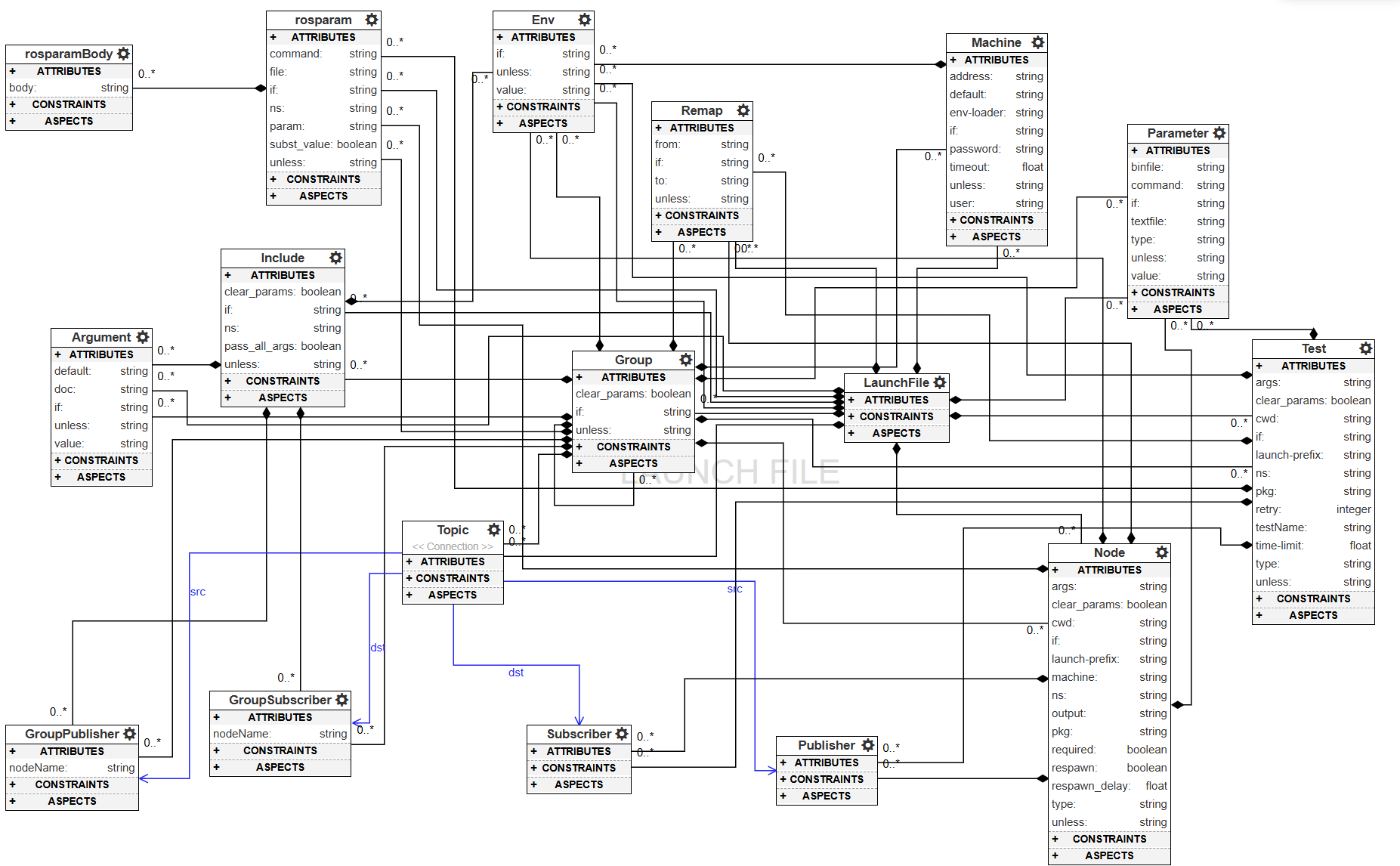}
    \caption{Metamodel}
    \label{fig:metamodel}
\end{figure}

The metamodel defines the structures and rules of the modeling language for ROSLaunchVisual. An image of this metamodel is shown in Figure \ref{fig:metamodel}. Each box represents a different component that can be added to the model. Most boxes have an exact match with a tag that can be added to an XML ROS launch file.

\begin{figure}
    \centering
    \includegraphics[width=0.3\linewidth]{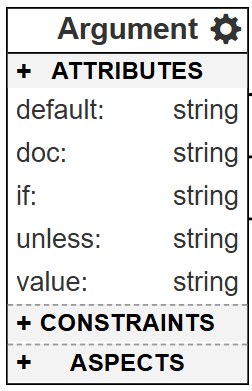}
    \caption{Example tag definition in metamodel}
    \label{fig:argument}
\end{figure}

A closer view of the \texttt{Argument} element is shown in Figure \ref{fig:argument}. The attributes listed in the box match the valid attributes of an \lstinline|<arg>| tag in a ROS launch file. The \texttt{name} attribute is added by default to all metamodel elements in WebGME, so it is not visible in this image.

The relationships between elements in the metamodel are represented by the lines that connect the different elements. The black lines with the diamonds on the end represent containment. The element touching the diamond contains the other element. For instance, the \texttt{Argument} tag can be contained inside a \texttt{LaunchFile} element, an \texttt{Include} element, or a \texttt{Group} element.

\begin{figure}
    \centering
    \includegraphics[width=\linewidth]{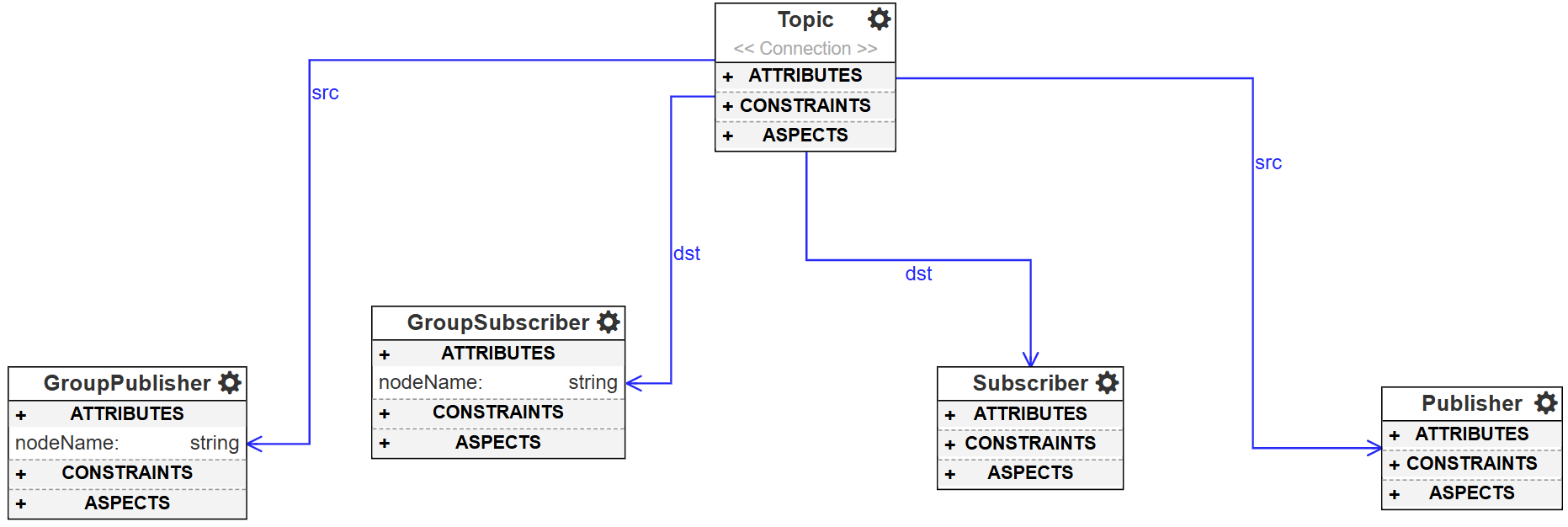}
    \caption{Topic connections definition in metamodel}
    \label{fig:connections}
\end{figure}

The other relationship present in the metamodel is a pointer relationship, shown with blue arrows. A closer view of the pointer relationship is shown in Figure \ref{fig:connections}. The pointer relationship allows connections to be drawn between elements in the model. In this metamodel, the \texttt{Topic} is the connection element. In ROS, a topic connects a publisher to a subscriber. This relationship is not bidirectional. Therefore, the \texttt{Publisher} elements are specified as the sources of the \texttt{Topic} connections, and the \texttt{Subscriber} elements are specified as the destinations for the \texttt{Topic} connections. It is important to note that the publisher and subscriber information is not present in an XML launch file. These elements will not be part of the generated output file. However, these elements must be included in the metamodel so that the communication can be visualized in the model during the design process.

The metamodel also defines other attributes of the elements in the model. For instance, the \texttt{Publisher} and \texttt{Subscriber} are designated as ports. Also, the default image and the displayed name of the elements in the model can be changed within the metamodel. These were changed for each element accordingly to enhance the overall user experience in ROSLaunchVisual. Furthermore, the metamodel contains extra sheets to store the library elements. These are predefined nodes, tests, and included launch files present in ROS code that a user may wish to add. The functionality of the library is discussed in more detail in Section \ref{section:update_lib}.

\subsection{Building and Editing a Launch File}

\begin{figure}
    \centering
    \includegraphics[width=\linewidth]{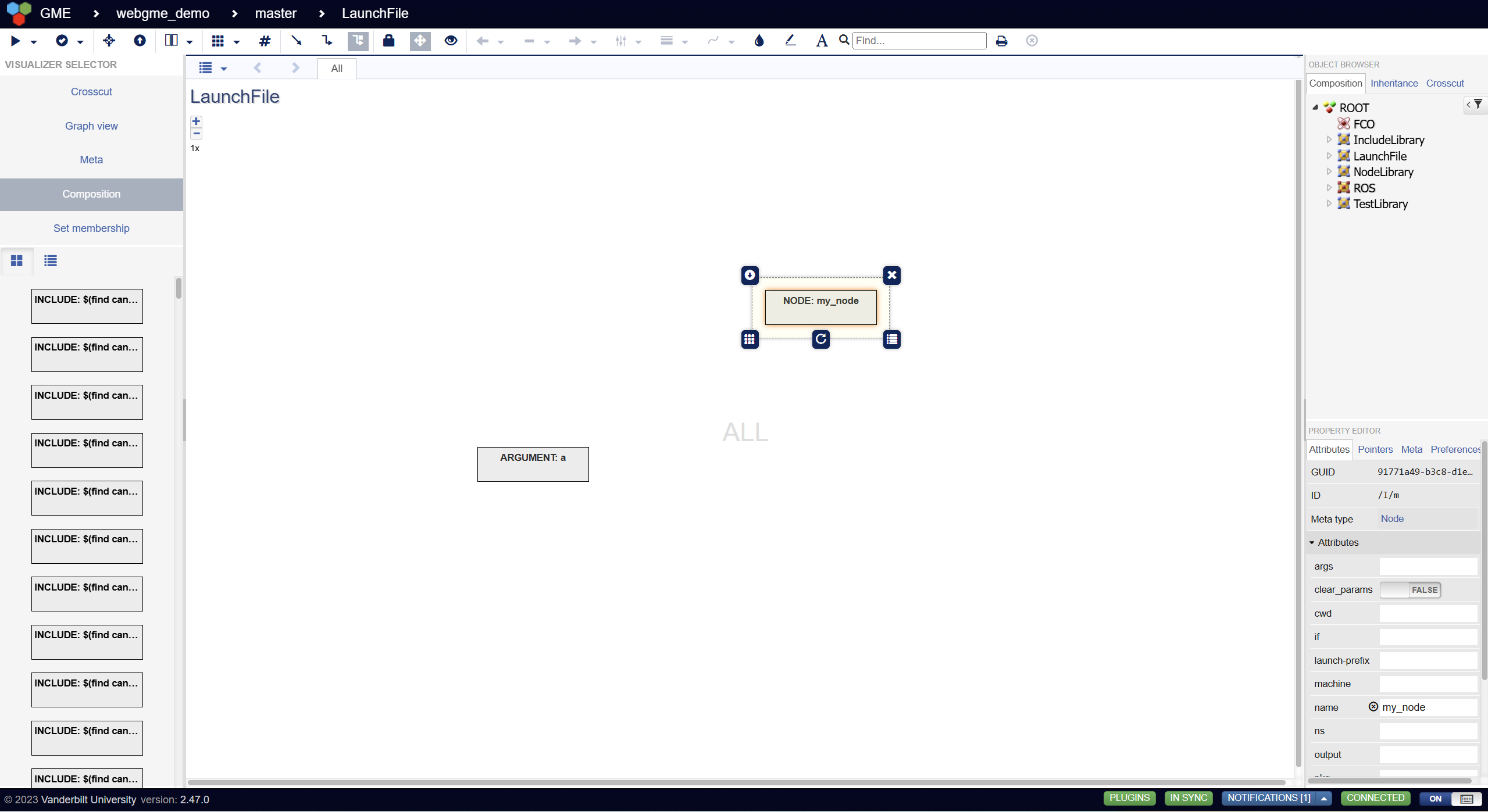}
    \caption{WebGME display}
    \label{fig:webgme-example}
\end{figure}

Upon launching WebGME, users select to continue working on an existing project or create a new project. New projects should be created with the ROSLaunch seed. At that point, the new project will be opened to the root level. The library folders are visible at this level. The user can then add a new launch file to the root of the project. Each launch file element in the model is representative of one file. The user can then descend into the model and begin working on the model, as shown in Figure \ref{fig:webgme-example}. New elements can be dragged and dropped from the left side of the screen. The attributes can be edited on the right. Traversing down into a node allows the user to update nested tags within an element.

\subsection{ROS Configuration Generator} \label{section:config-gen}
In an XML launch file, there is almost no information about the publishers and subscribers that are available on the nodes. There may be a remap tag nested inside a node tag, implying that the node has a publisher or subscriber with that name present and that the publisher or subscriber name should be changed for the purpose of running this launch file. Other than this, the information about publishers and subscribers cannot be retrieved from the XML file. It must be extracted from the source code of the nodes. The ROS Configuration Generator is used to parse this information from several packages so that communication between publishers and subscribers can be observed in ROSLaunchVisual.

\begin{figure}
    \centering
    \includegraphics[width=0.5\linewidth]{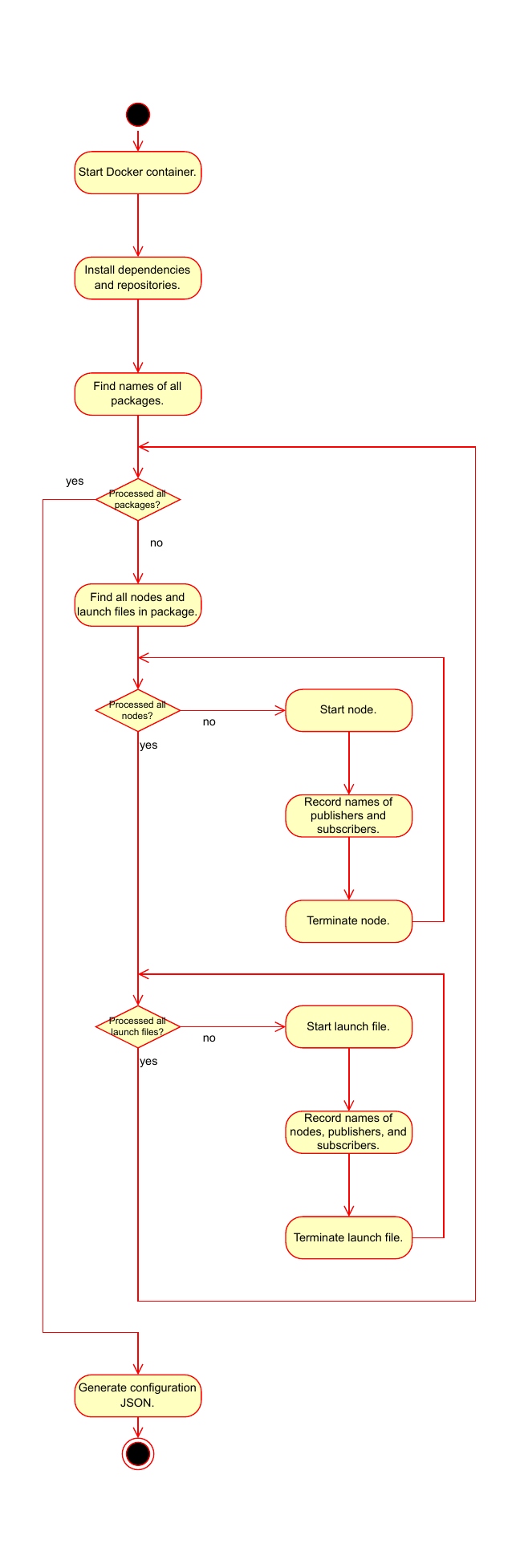}
    \caption{ROS Configuration Generator process}
    \label{fig:ros-config-generator-diagram}
\end{figure}

The ROS Configuration Generator is a Docker container that analyzes ROS repositories hosted on GitHub and outputs a list of all the publishers and subscribers on all the nodes and launch files present in every package. The input to the tool is a YAML file that specifies the owner and name of each repository and, if necessary, the desired branch to check out. The input file can also include any packages or libraries installed with pip or apt-get that the ROS packages require. The container is built on top of a ROS image. A diagram of the process of running the Docker container is shown in Figure \ref{fig:ros-config-generator-diagram}. Upon starting the container, all the dependencies listed in the input file are installed, and all the ROS repositories are cloned into a ROS workspace. The container then installs all the dependencies in the ROS workspace. The next step is to identify all the ROS packages that have been installed. The tool then analyzes each package. For a package, it identifies all the nodes and the launch files that are present. For each node, the tool starts the node, waits two seconds, captures all of the publishers and subscribers present, and then terminates the node. For each launch file, the tool launches the launch file, waits two seconds, captures all running nodes and their associated publishers and subscribers, and then terminates the launch file. All of this information for every package is collected as a JSON and produced as the output file. This file can then be used to build a library for ROSLaunchVisual (as described further in Section \ref{section:update_lib}), allowing the tool to use information about node communication that is not present in a launch file. An evaluation of the ROS Configuration Generator is presented in Section \ref{section:config_gen_eval}.

\subsection{Plugins}
WebGME allows developers to build plugins to enhance the user model development experience. WebGME can automatically generate a plugin skeleton for Python or JavaScript code. Five plugins were developed for ROSLaunchVisual, and each is discussed in more detail below. Due to the highly flexible nature of WebGME, it would be simple for anyone to build their own plugin to add to these five and extend the capabilities of ROSLaunchVisual.

\subsubsection{Update Library} \label{section:update_lib}
In modeling a ROS launch file, the first plugin to use is the Update Library plugin. This takes the configuration generated by the ROS Configuration Generator (see Section \ref{section:config-gen}) and builds up a library of nodes, test nodes, and included launch files with their publishers and subscribers predefined.

Upon execution, the plugin parses the input JSON and creates new nodes. These are also added to the \texttt{Node Library} page, \texttt{Test Library} page, or \texttt{Include Library} page in the metamodel. By adding these to the metamodel, the library elements are prepopulated in the left panel for users to drag and drop into the launch file. This makes it easy to add existing ROS components to a launch file without having to search a ROS workspace and redefine the information manually.

\subsubsection{Import Launch}
The Import Launch plugin ingests an existing ROS~1 XML launch file and reconstructs its structure in the WebGME environment. Upon execution, the plugin locates the file specified in the project configuration and parses its contents using Python’s \lstinline|xml.etree.ElementTree| library. Each launch element is mapped to a metamodel construct: tags are normalized (e.g., \lstinline|<arg>|→\texttt{Argument}, \lstinline|<param>|→\texttt{Parameter}), and special-case attributes such as an \lstinline|include| tag’s \lstinline|file| or a \lstinline|group| tag’s \lstinline|ns| are renamed to ``\lstinline|name|''. The result is a nested dictionary that preserves tag hierarchies, attributes, and child relationships.

Once parsing is complete, the plugin ensures that the \texttt{LaunchFile} metatype exists before creating a new root node under the active project folder. It then traverses the parsed hierarchy recursively: for each node or test element, the plugin attempts to match its \lstinline|pkg| and \lstinline|type| attributes against entries in the prepopulated \texttt{Node Library} or \texttt{Test Library}. For include elements, the plugin attempts to match the \lstinline|name| attribute. If a match exists, the library node is copied along with its static attributes and any nested \texttt{Publisher} or \texttt{Subscriber} children, preserving known topic names. Otherwise, a fresh modeling element of the corresponding metatype is instantiated and all XML-specified attributes are applied (with \lstinline|true|/\lstinline|false| converted to booleans). After instantiating all elements, the plugin commits the new subtree to the active branch, producing a visual model that accurately reflects the original launch file while incorporating publisher and subscriber information from the library.

\subsubsection{Make Connections}
To see the connections between nodes in a launch file, the Make Connections plugin can be used. The plugin connects the publishers and subscribers with the same name (or those that have been remapped to have the same name) using arrows. The arrows have a source at the publisher and a destination at the subscriber. The name of the topic used to communicate is displayed on the arrow. The \texttt{GroupPublisher} and \texttt{GroupSubscriber} elements of the metamodel allow connections between nodes nested inside groups or included launch files to be displayed.

The plugin execution begins by deleting any existing \texttt{Topic} connections and \texttt{GroupPublishers} and \texttt{GroupSubscribers} that are the child of a \texttt{Group}. New \texttt{GroupPublishers} and \texttt{GroupSubscribers} are then added from the bottom up, with any namespaces and remaps being applied as specified by the model. After that, \texttt{Publisher} and \texttt{Subscriber} elements at the same level in the model are connected with an arrow, following the \texttt{Remap} elements that have been added to the model. At this point, all potential connections are displayed in the launch file. This makes it easy to see how publishers and subscribers would connect in the running launch file.

\subsubsection{Error Checking}
The Error Checking plugin checks for three potential errors in a launch file. These are checking for duplicate node names, checking for errors in argument tag definitions, and checking for circular dependencies in arguments. There are many other errors that could be checked, but for this tool, only three error checks are currently supported. A notification message with the final error report is displayed at the bottom of the screen once the plugin finishes running.

When checking for duplicate names, the model is parsed for all nodes and tests. Any duplicate names (after namespaces have been applied) are reported. In checking for errors with arguments, the attributes of the \texttt{Argument} elements are examined. In an argument, the default attribute or the value attribute can be provided, but not both. Arguments that are not defined correctly are listed in the error report. The final check is for circular argument substitution. Arguments can be defined as using the values of other arguments through substitution. However, circular dependencies where, for instance, argument A depends on the value of argument B and argument B depends on the value of argument A, will cause problems. The plugin ensures that no circular arguments are present in the model. If errors are found, they are added to the error report.

\subsubsection{Export Launch}
The Export Launch plugin is a core feature of ROSLaunchVisual, responsible for translating a graphical model into a valid ROS XML launch file. This transformation is not a simple one-to-one mapping; rather, it requires parsing a structured model hierarchy, interpreting user-defined attributes, resolving dependencies, and formatting the output according to ROS conventions.

The plugin starts by traversing the model’s root \texttt{LaunchFile} node and recursively processes its child components, such as \texttt{Node}, \texttt{Group}, \texttt{Argument}, and \texttt{Remap} elements. For each component, it extracts relevant attributes like package names, executable types, namespaces, and argument values, then formats them into proper XML tags. Special care is taken to handle argument substitution correctly: arguments that depend on other arguments are ordered to ensure valid substitution during runtime. Tags are also ordered to ensure precedence relationships are maintained. For instance, remap tags are output before the node definitions they affect.

One of the key advantages of this plugin is its ability to preserve model-level validations in the final launch file. Once the export is complete, users are provided with a downloadable XML file that can be directly used in a ROS system without any manual editing. This significantly shortens the development cycle and ensures the launch file faithfully reflects the graphical model. The Export Launch plugin thus serves as a crucial bridge between the visual design environment and the executable deployment artifacts in a ROS-based robotics workflow.

\section{Case Study} \label{section:casestudy}
This section covers a case study of using the tool to edit the launch file shown in Listing \ref{code:input}. The goal of the launch file is to launch four turtlesim nodes from the turtlesim package in the ROS tutorials \cite{ros_tutorials}. The controllers for these turtlesim nodes come from the tutorial package and a custom package. The controllers are designed to drive the turtles in a circle, a square, a straight line, and  a circle with a radius value specified by an argument. The current version of the launch file has several issues that need to be addressed.

\FloatBarrier
\begin{samepage}
\lstset{
    language=xml,
    frame=single,
    framexleftmargin=5pt,
    framexrightmargin=5pt,
    framesep=12pt,
    tabsize=3,
    caption=input.launch,
    label=code:input,
    rulesepcolor=\color{gray},
    commentstyle=\color{OliveGreen},
    stringstyle=\color{red},
    numbers=left,
    numberstyle=\tiny,
    numbersep=5pt,
    breaklines=true,
    showstringspaces=false,
    basicstyle=\ttfamily\footnotesize,
    emph={},emphstyle={\color{magenta}},
}
\begin{lstlisting}
<launch>
    <arg name="radius" default="3.0"/>

    <node pkg="turtlesim" name="sim1" type="turtlesim_node"/>
    <node pkg="my_turtle_pkg" name="circle" type="turtle_circle"/>

    <node pkg="turtlesim" name="sim2" type="turtlesim_node"/>
    <node pkg="turtlesim" name="square" type="draw_square"/>

    <node pkg="turtlesim" name="sim2" type="turtlesim_node">
        <remap from="turtle1/cmd_vel" to="line_cmd_vel"/>
    </node>
    <node pkg="my_turtle_pkg" name="line" type="turtle_line"/>
        
    
    <node pkg="turtlesim" name="sim4" type="turtlesim_node"/>
    <node pkg="my_turtle_pkg" name="circle_radius" type="turtle_circle_radius" args="$(arg radius)"/>
</launch>
\end{lstlisting}
\end{samepage}
\FloatBarrier

\begin{figure}[htbp]
    \centering
    \begin{subfigure}{\textwidth}
        \centering
        \includegraphics[width=0.9\textwidth]{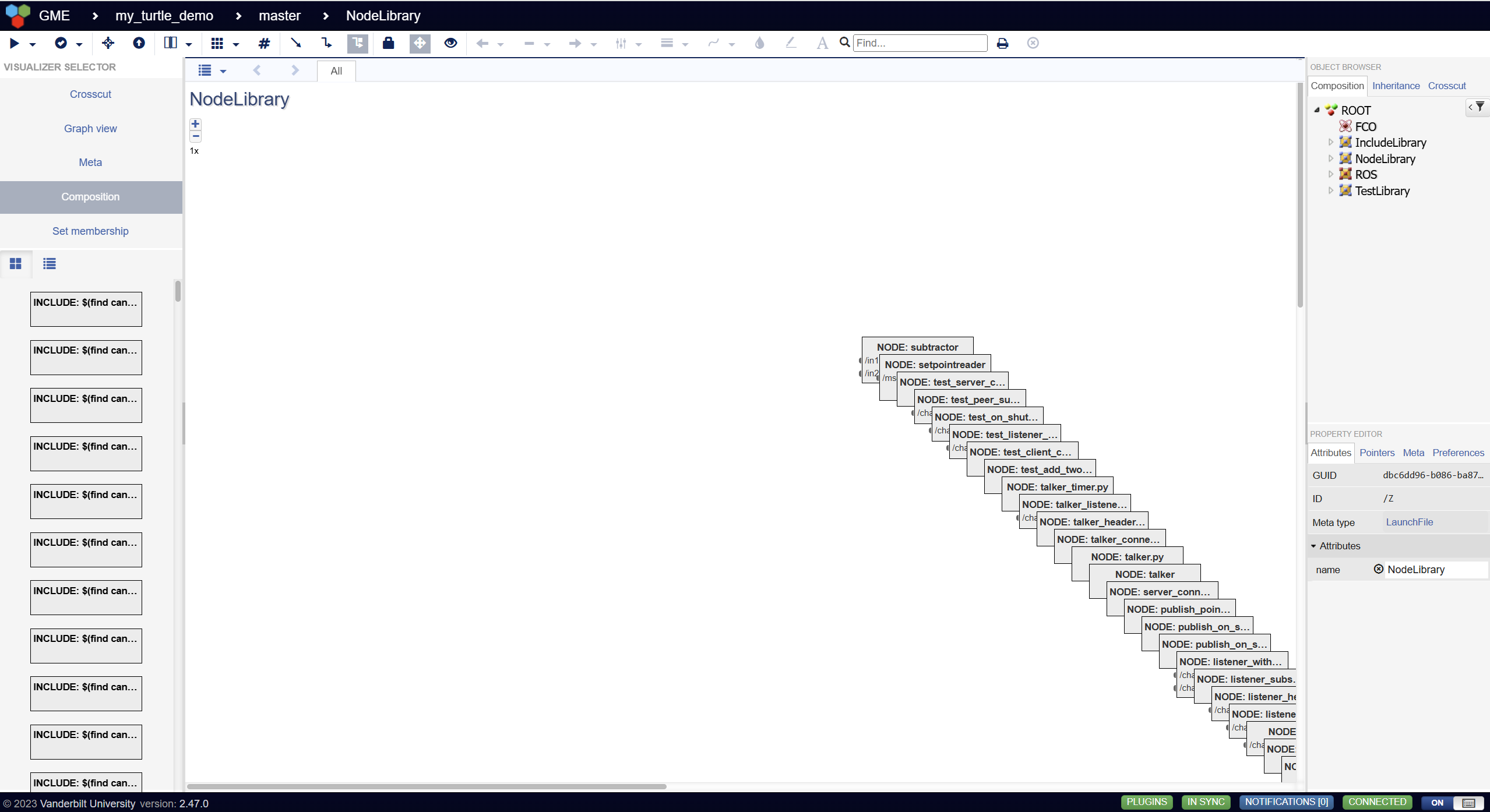}
        \subcaption{Before update}
        \label{fig:lib-before}
    \end{subfigure}

    \begin{subfigure}{0.9\textwidth}
        \centering
        \includegraphics[width=\textwidth]{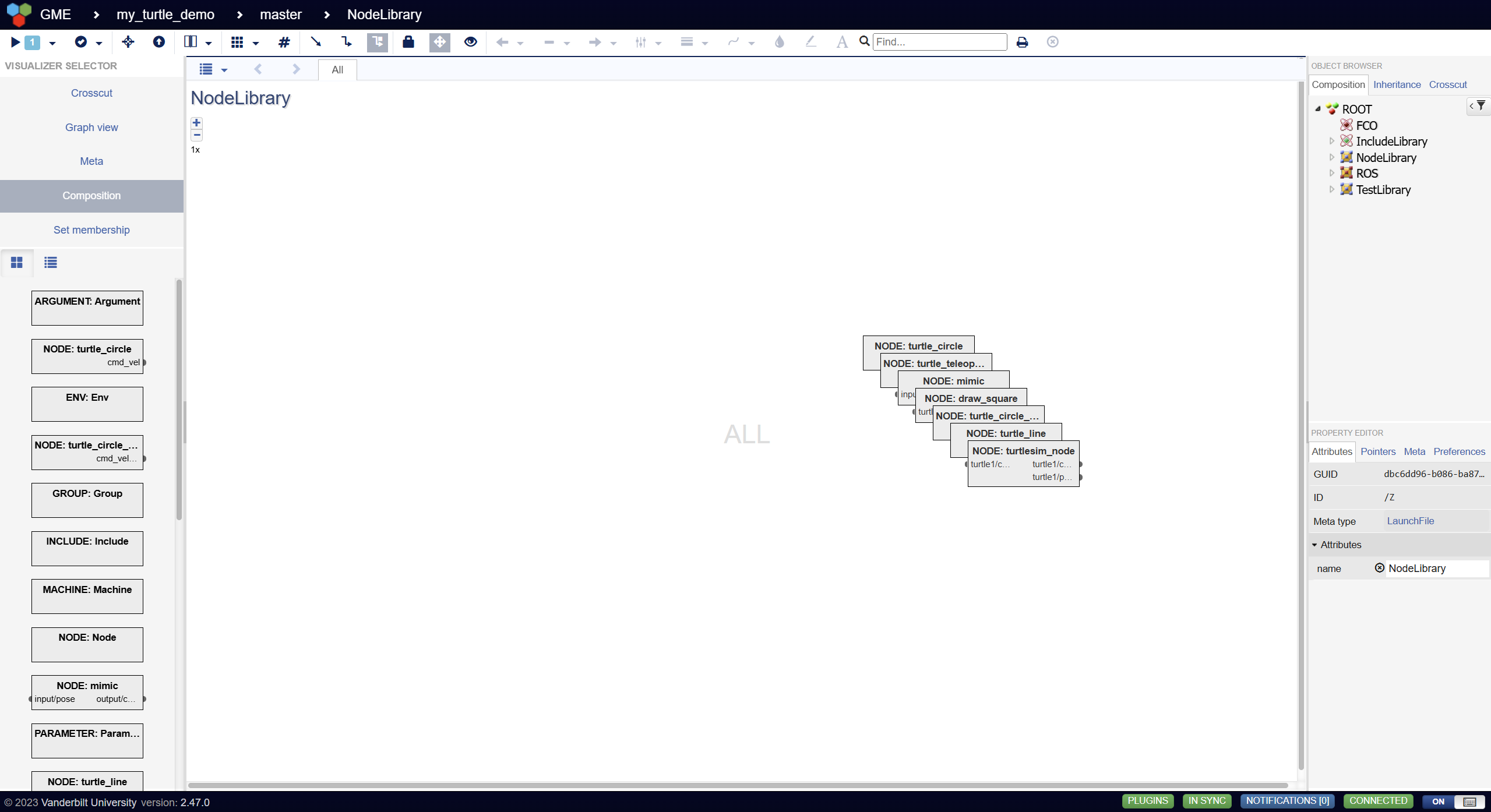}
        \subcaption{After update}
        \label{fig:lib-after}
    \end{subfigure}
    
    \caption{Comparison of the node library before and after the update}
    \label{fig:updatelib-before-after}
\end{figure}

The first step is to update the library to include nodes found in the necessary packages. A configuration JSON was generated for the packages used in the launch file. Corrections to the JSON were made to adjust for limitations of the ROS Configuration Generator (see Section \ref{section:config_gen_eval} for an explanation of why this is necessary). The configuration JSON was then provided to the Update Library plugin to add the new nodes for use in the model. Figure \ref{fig:updatelib-before-after} shows the node library before and after plugin execution.

\begin{figure}
    \centering
    \includegraphics[width=\linewidth]{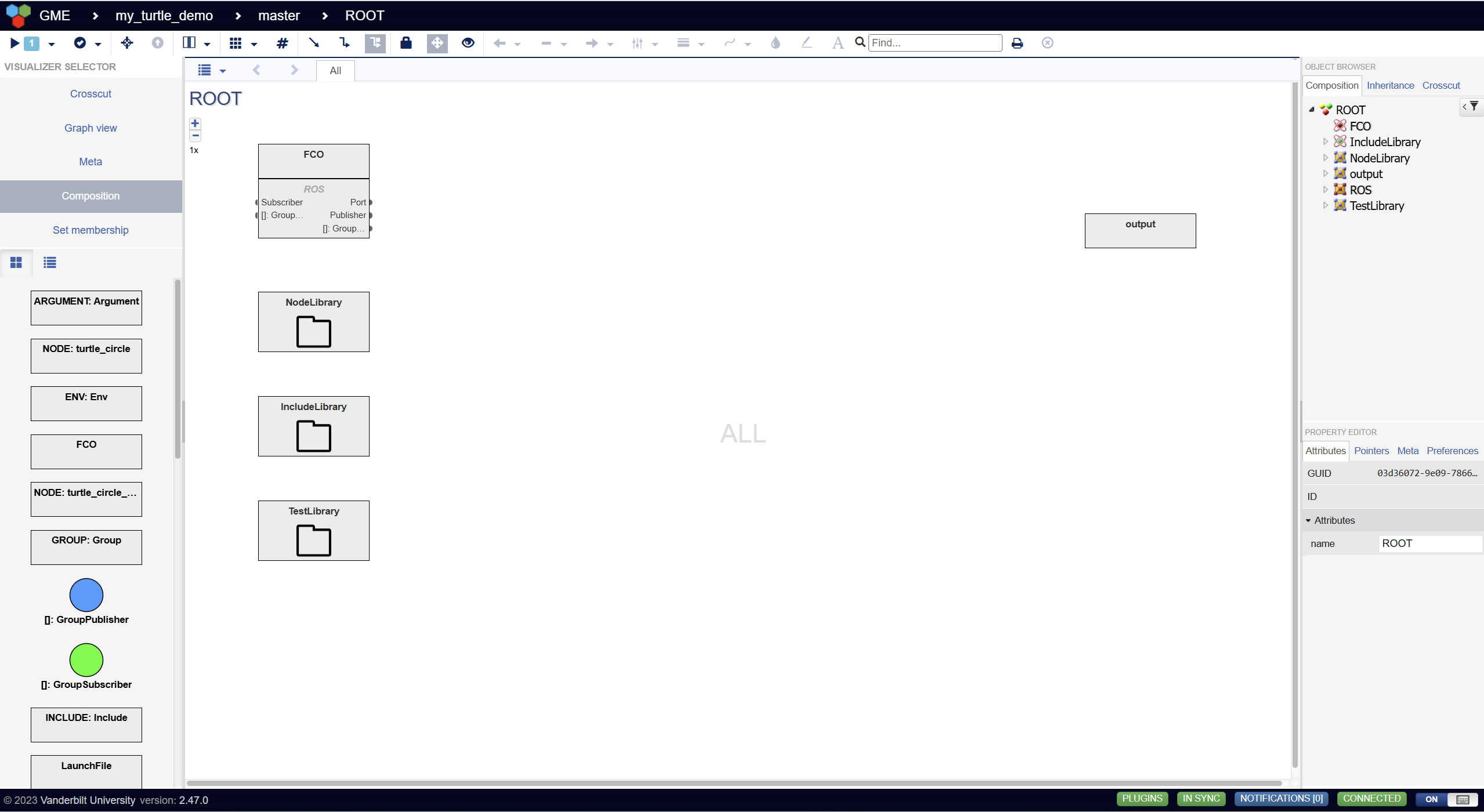}
    \caption{Result of importing launch file}
    \label{fig:input-launch-result}
\end{figure}

The next step is to upload the input.launch file to the tool. Upon execution of the Import Launch plugin, a new file is added to WebGME, as shown in Figure \ref{fig:input-launch-result}. The name should be changed to the desired output file name.

\begin{figure}
    \centering
    \includegraphics[width=\linewidth]{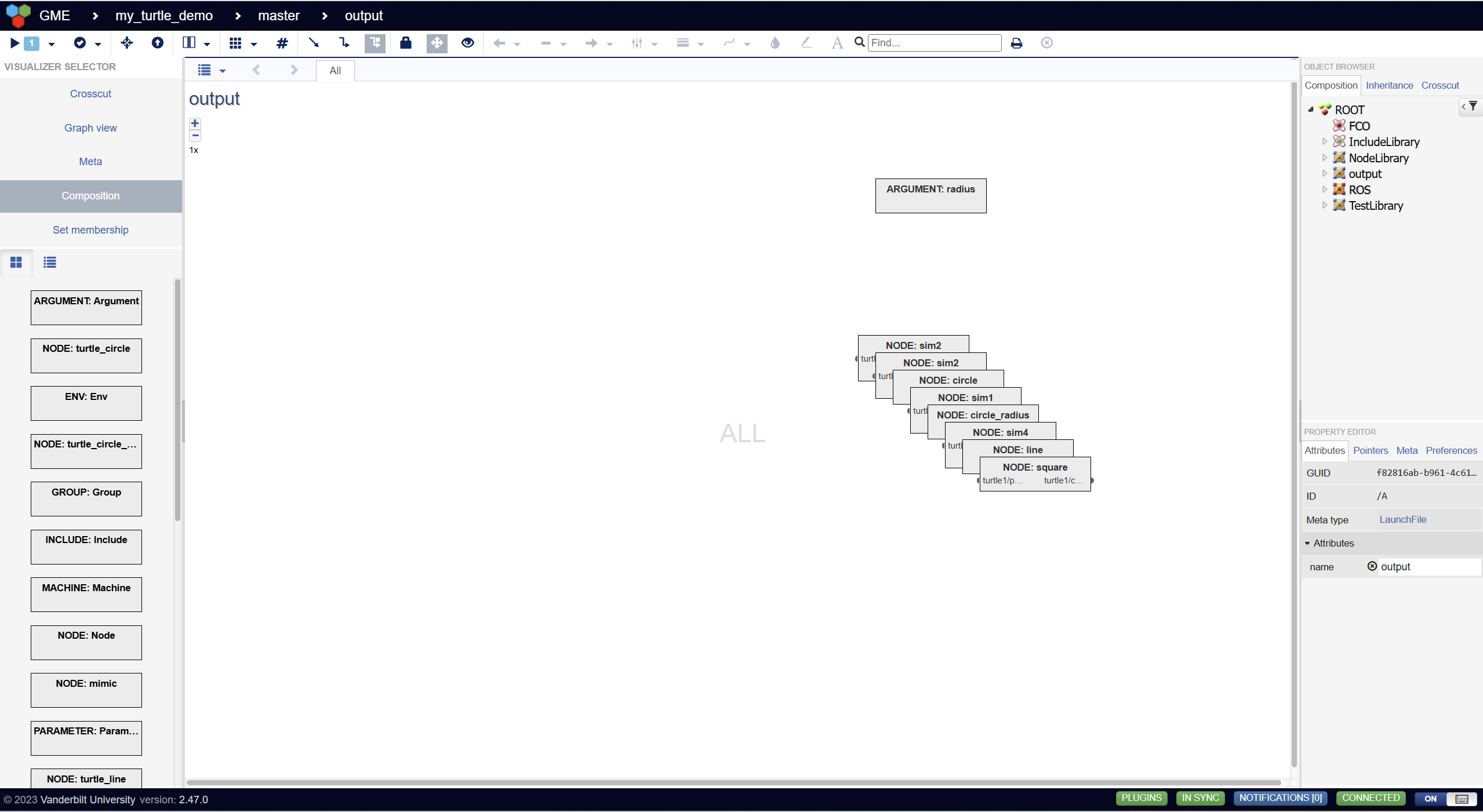}
    \caption{Top level of imported launch file}
    \label{fig:import-top}
\end{figure}

Inside the launch file model at the top level, all top-level tags are visible, as shown in Figure \ref{fig:import-top}.

\begin{figure}
    \centering
    \includegraphics[width=\linewidth]{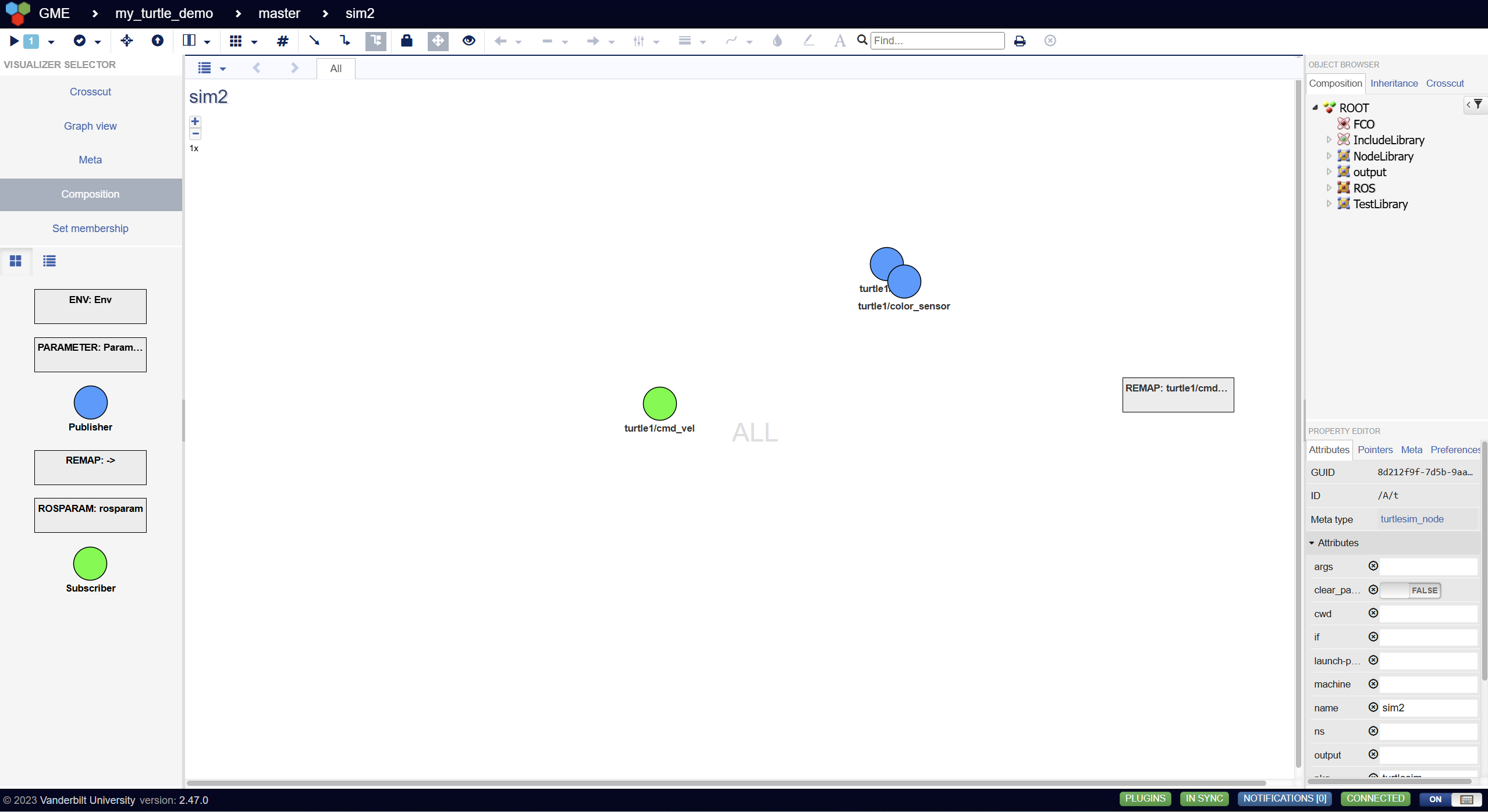}
    \caption{Inside the sim2 node tag}
    \label{fig:in-sim2}
\end{figure}

The nested tags are also included in the model. Inside the sim2 node, the remap tag is present, as shown in Figure \ref{fig:in-sim2}. The publishers and subscriber are also visible from inside the sim2 node.

\begin{figure}
    \centering
    \includegraphics[width=\linewidth]{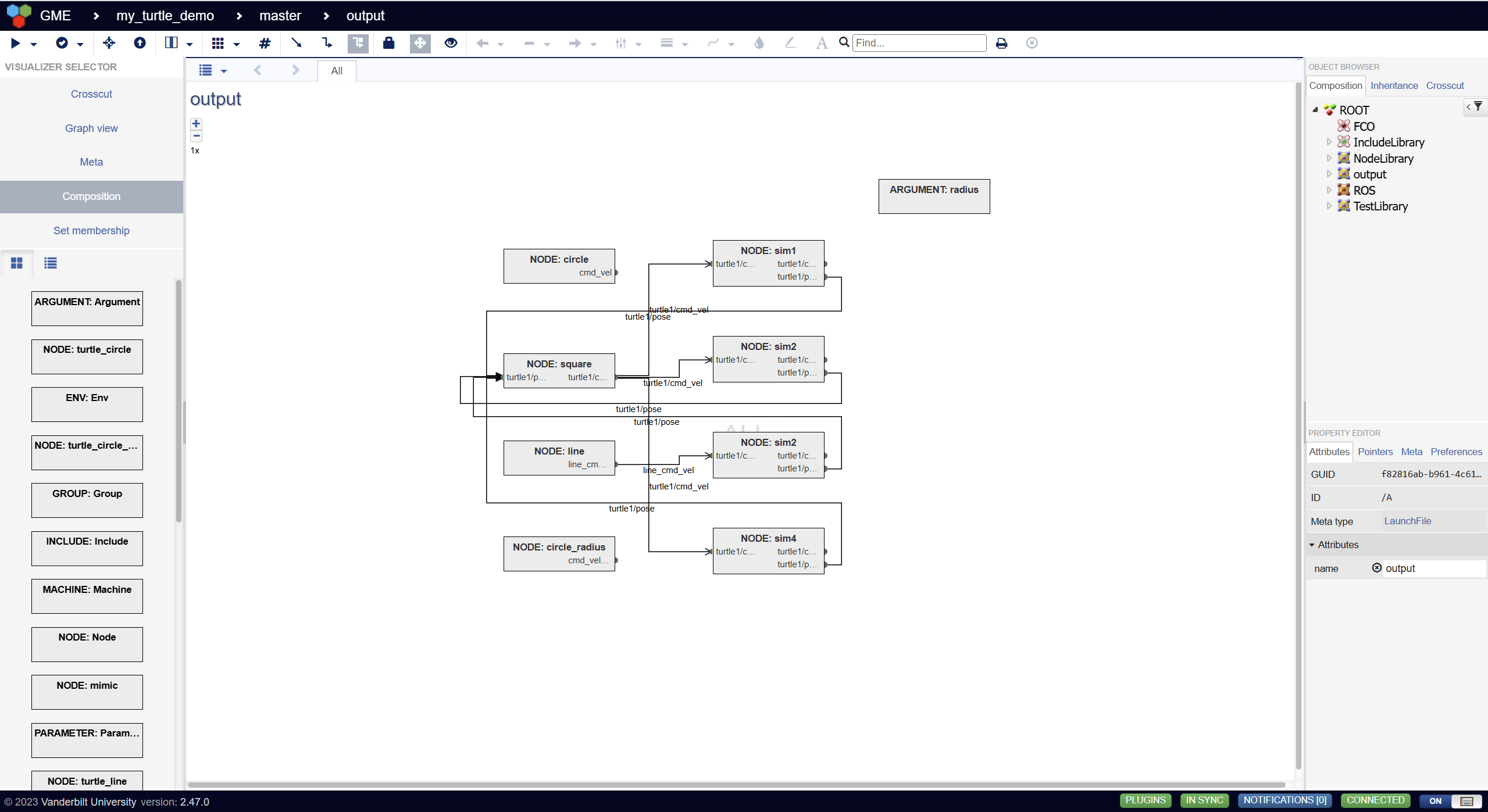}
    \caption{Result of Make Connections plugin execution on original file}
    \label{fig:bad-make-connections}
\end{figure}

The next step is to begin using the Make Connections and Error Checking plugins to fix the issues with the file. Running the Error Checking plugin reveals that there are two nodes named sim2, which is not allowed. The results of running the Make Connections plugin, shown in Figure \ref{fig:bad-make-connections}, also reveal problems. The topic names have not been appropriately remapped. The controllers are not communicating with the appropriate turtlesims. These results reveal some insight into what is currently going wrong and what changes need to be made.

\begin{figure}[htbp]
    \centering
    \includegraphics[width=\textwidth]{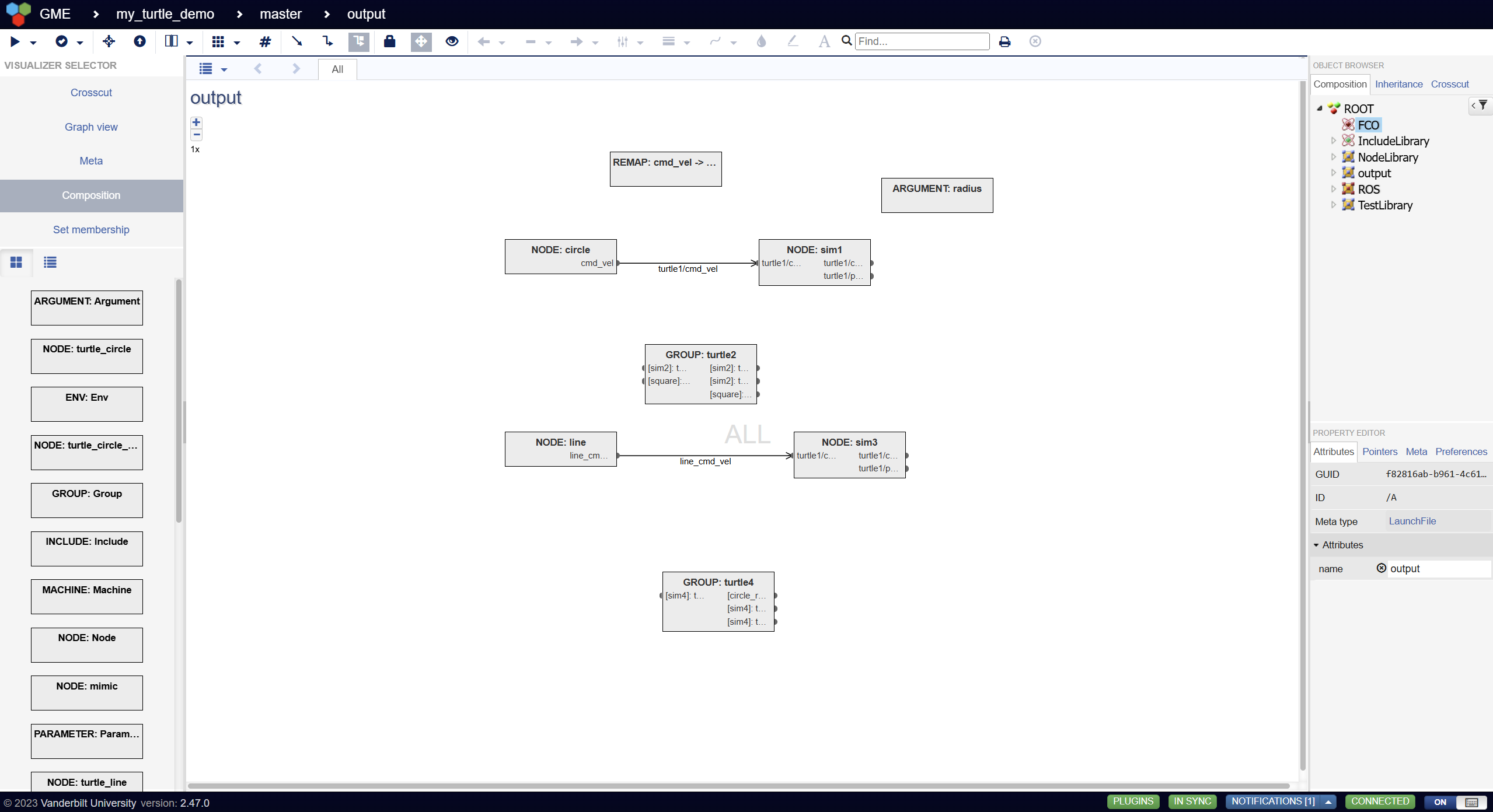}
    \caption{Result of Make Connections plugin at the top level.}
    \label{fig:top-connect}
\end{figure}

\begin{figure}[htbp]
    \centering
    \includegraphics[width=\textwidth]{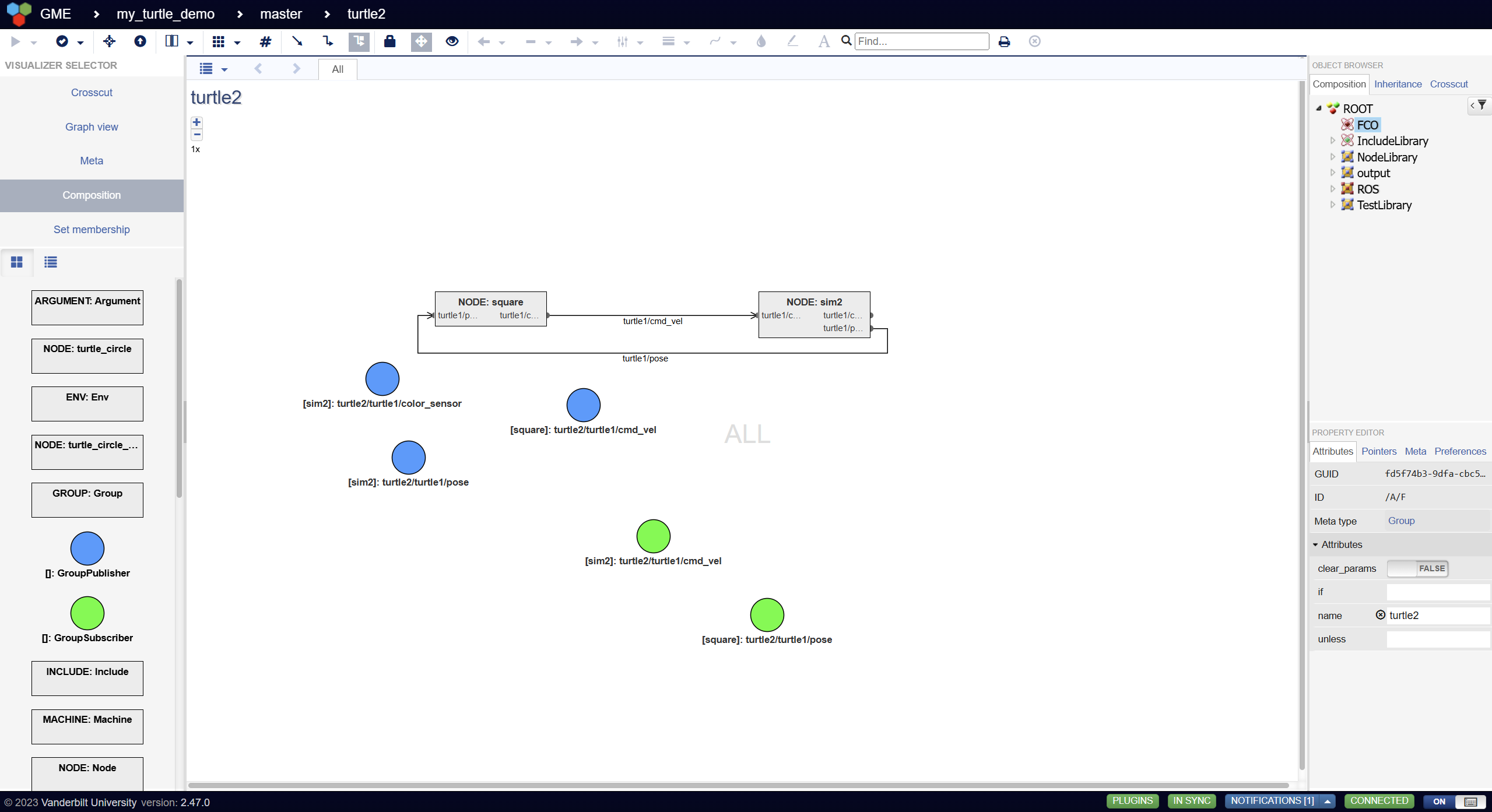}
    \caption{Result of Make Connections plugin inside the turtle2 group.}
    \label{fig:turtle2-connect}
\end{figure}

\begin{figure}[htbp]
    \centering
    \includegraphics[width=\textwidth]{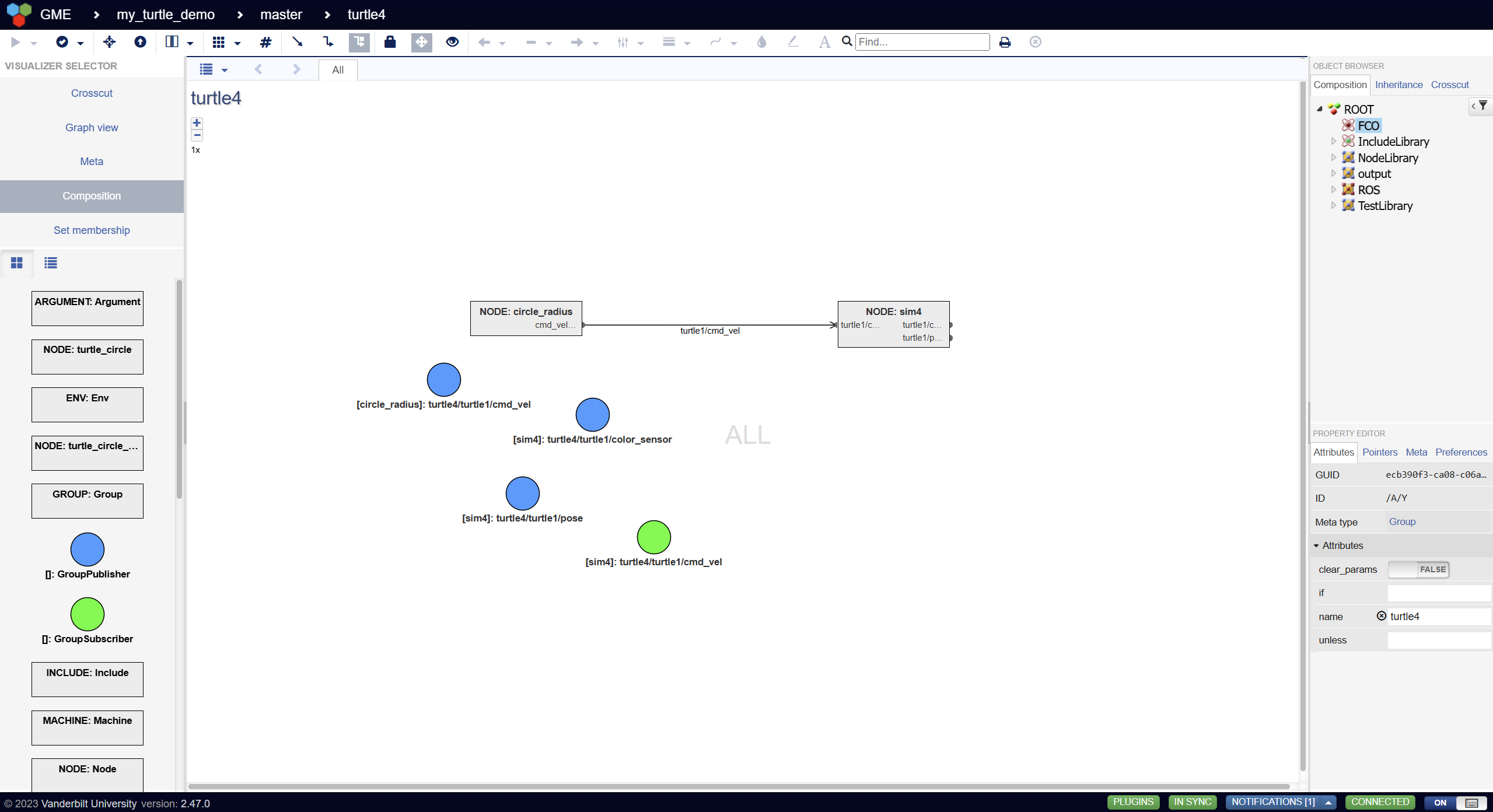}
    \caption{Result of Make Connections plugin inside the turtle4 group.}
    \label{fig:turtle4-connect}
\end{figure}

After adding some remap and group tags and fixing the duplicate node name, the controllers are now connected to the appropriate turtlesim nodes. The result of running the Make Connections plugin again is shown in Figures \ref{fig:top-connect}, \ref{fig:turtle2-connect}, and \ref{fig:turtle4-connect}.

\begin{figure}
    \centering
    \includegraphics[width=\linewidth]{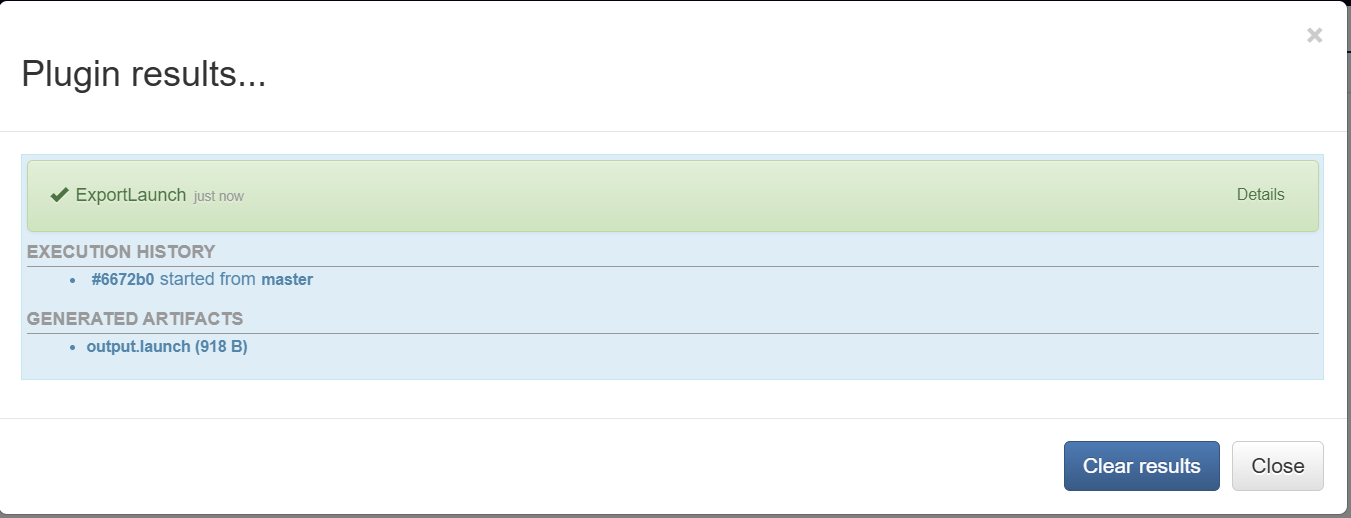}
    \caption{Result of Export Launch plugin}
    \label{fig:export-result}
\end{figure}

Now that there are no more errors and everything communicates as expected, the final model can be exported into the XML format. After running the Export Launch plugin, clicking the link in the result will download the file (see Figure \ref{fig:export-result}). At this point, the file can be moved to the appropriate location to be executed.

The output file is shown in Listing \ref{code:output}. All tags have been written to the file with the correct hierarchy. The turtlesim simulation in the launch file will now run as expected.

\FloatBarrier
\begin{samepage}
\lstset{
    language=xml,
    frame=single,
    framexleftmargin=5pt,
    framexrightmargin=5pt,
    framesep=12pt,
    tabsize=3,
    caption=output.launch,
    label=code:output,
    rulesepcolor=\color{gray},
    commentstyle=\color{OliveGreen},
    stringstyle=\color{red},
    numbers=left,
    numberstyle=\tiny,
    numbersep=5pt,
    breaklines=true,
    showstringspaces=false,
    basicstyle=\ttfamily\footnotesize,
    emph={},emphstyle={\color{magenta}},
}
\begin{lstlisting}
<launch>
  <arg name="radius" default="3.0"/>
  <remap from="cmd_vel" to="turtle1/cmd_vel"/>
  <group ns="turtle4">
      <node name="circle_radius" pkg="my_turtle_pkg" type="turtle_circle_radius" args="$(arg radius)">
          <remap from="cmd_vel_radius" to="turtle1/cmd_vel"/>
      </node>
      <node name="sim4" pkg="turtlesim" type="turtlesim_node">
      </node>
  </group>
  <group ns="turtle2">
      <node name="square" pkg="turtlesim" type="draw_square">
      </node>
      <node name="sim2" pkg="turtlesim" type="turtlesim_node">
      </node>
  </group>
  <node name="line" pkg="my_turtle_pkg" type="turtle_line">
  </node>
  <node name="sim1" pkg="turtlesim" type="turtlesim_node">
  </node>
  <node name="circle" pkg="my_turtle_pkg" type="turtle_circle">
  </node>
  <node name="sim3" pkg="turtlesim" type="turtlesim_node">
      <remap from="turtle1/cmd_vel" to="line_cmd_vel"/>
  </node>
</launch>
\end{lstlisting}
\end{samepage}
\FloatBarrier

\section{Evaluation} \label{section:evaluation}

The following section describes the evaluation of two components of the tool. These are the ROS Configuration Generator and the import and export of launch file. Both pieces of the tool are tested with real ROS code and evaluated for areas that perform well and which use cases the tool cannot fully handle. The runtime of all plugin execution is also evaluated.

\subsection{ROS Configuration Generator} \label{section:config_gen_eval}

The performance of the ROS Configuration Generator was evaluated on several ROS packages that are used by this research group and that are commonly available to the research community. These packages \cite{ros_comm, geometry2, moveit, ros_diagnostics, 10.1145/3459609.3460531, jmscsl_cbf_repo, jmscsl_velocity_controller, jmscsl_setpointreader, jmscsl_oneshotadder, jmscsl_clock2seconds, jmscsl_odometer, jmscsl_modechanger, jmscsl_subtractor, matthewnice_joy_to_veh, matthewnice_joystick_drivers, jmscsl_profproject, nice2023sailing, ros_tutorials} were run through the configuration process individually and collectively to evaluate the nodes and launch files. Table \ref{tab:package_summary} provides a count of launch files available in each repository as identified by the tool. The build logs, run logs, and JSON output configuration were analyzed. In this section, a summary of the major findings of the analysis is presented.

\begin{table}[htbp]
\caption{Summary of ROS Repositories}
\label{tab:package_summary}
\begin{tabular}{@{}ll@{}}
\toprule
Repository & Count of Launch Files \\
\midrule
\texttt{can\_to\_ros} \cite{10.1145/3459609.3460531} & 56 \\
\texttt{cbf} \cite{jmscsl_cbf_repo} & 0 \\
\texttt{clock2seconds} \cite{jmscsl_clock2seconds} & 0 \\
\texttt{geometry2} \cite{geometry2} & 11 \\
\texttt{gps2vsl} \cite{nice2023sailing} & 3 \\
\texttt{joy\_to\_veh} \cite{matthewnice_joy_to_veh} & 2 \\
\texttt{joystick\_drivers} \cite{matthewnice_joystick_drivers} & 2 \\
\texttt{modechanger} \cite{jmscsl_modechanger} & 0 \\
\texttt{moveit} \cite{moveit} & 27 \\
\texttt{odometer} \cite{jmscsl_odometer} & 0 \\
\texttt{oneshotadder} \cite{jmscsl_oneshotadder} & 0 \\
\texttt{profproject} \cite{jmscsl_profproject} & 1 \\
\texttt{ros\_diagnostics} \cite{ros_diagnostics} & 23 \\
\texttt{ros\_tutorials} \cite{ros_tutorials} & 19 \\
\texttt{roscomm} \cite{ros_comm} & 18 \\
\texttt{setpointreader} \cite{jmscsl_setpointreader} & 0 \\
\texttt{subtractor} \cite{jmscsl_subtractor} & 0 \\
\texttt{velocity\_controller} \cite{jmscsl_velocity_controller} & 0 \\
\bottomrule
\end{tabular}
\end{table}

It was observed that the ROS Configuration Generator does not always correctly identify all of the nodes that are present in a package. 
In some cases, the ROS Configuration Generator identified other executables that are not ROS nodes as ROS nodes. The logic in the program for identifying nodes mimics the logic found in the autocomplete functionality of the command line command ``rosrun package\_name node\_name''. Given a package name in this command line command, when the user presses tab, a list of all nodes in the package will be provided. This logic appears to return any executable that is found within a certain file path. Sometimes this feature (native to ROS) returns executables that are not actually ROS nodes. For the purposes of this tool, it was determined that it was acceptable if extra executables were listed as nodes. In these extra executables, no publishers or subscribers are identified, which is a clue that these may not be real nodes. The user would simply need to remove these from the output configuration file before loading the library. On the other hand, the user could also simply choose to focus on the nodes in the library that they need for the launch file, ignoring superfluous false nodes.

In other cases, the program failed to identify all nodes in the packages. Upon further investigation, it was discovered that the CMAKE file would only build these missing nodes if the correct hardware is detected. Since the Docker container was not the intended platform for running these nodes and did not have this hardware, the nodes were not built. This is a limitation of running the nodes in a way that was not originally intended. 

In running the nodes, several error messages were generated about the inability to locate a particular hardware component. A common issue was the lack of a display in the Docker container. Since Docker is purely software, it is difficult to perfectly run code that is designed to be run on hardware. This problem may lead to the failure to correctly identify any or all of the publishers and subscribers in a given package. 

Another common issue discovered in this analysis was the inability of the Docker container to infer the correct command line arguments that should be used when starting a node or a launch file. Many of the nodes and launch files could not run without the correct arguments. However, it is difficult for the script to infer that the program needs command-line arguments, how many arguments the program needs, and what types of arguments are required. These nodes and launch files often crashed and did not return any of the publishers and subscribers present in the nodes.

Some of the nodes also expected to find certain files present on the system, such as parameters or other configuration options. If this was not present in the GitHub repository, then it would be highly unlikely that this resource would be available in the Docker image. This also would lead to an empty list of publishers and subscribers present in a node or launch file.

In some nodes, the publishers and subscribers present may be conditional. For instance, the subscribers on a rostopic echo node can be anything depending on what arguments are specified. It would be difficult to design the program to anticipate how different conditions may affect node behavior and then replicate these conditions to fully capture all potential node behavior.

The presence of other nodes may interact and change the publishers and subscribers present. This behavior was observed when running multiple packages together through the ROS Configuration Generator. Different publishers and subscribers were present than when one package was run at a time. When using the ROS Configuration Generator, it may be beneficial to include only packages that will be run together in the intended environment.

The output configuration lists all topics with a / at the beginning, which indicates that they are all in the global namespace. When running the ROS Configuration Generator, the tool queries the running topics and receives the fully resolved global names. Therefore, for a given node, the tool cannot distinguish whether a topic originally specified a relative or global topic name. The user is responsible for making the appropriate adjustments after the tool is run.

Also, when running multiple packages in the ROS Configuration Generator, launch files and nodes would not always terminate cleanly. This led to the result showing some launch files including nodes that are not part of the launch file. Instead, they were residuals from a previous execution that did not end cleanly. This is a limitation of execution within a Docker container.

For the cases where the current version of the ROS Configuration Generator does not perform well, there are options to improve performance. One would be editing the Dockerfile to better match the desired use case. The current program is designed to handle a variety of general cases. Tailoring the Docker image to more closely match a specific use case may lead to improved results. In addition, the output configuration file can be manually edited to correct for unexpected results. Then, the updated file can be loaded into the visual language to load the most accurate library.
Running the script on the intended hardware could also achieve more accurate results. However, it is important to do so safely. The ROS Configuration Generator starts and stops all nodes on the machine. Care should be taken to ensure that no people or equipment will be harmed in the process of running all the software one node at a time.

\subsection{Import and Export of Launch Files}

The import and export of thirty-six different launch files were tested to ensure that no changes were made in the process. These launch files came from the roscpp\_tutorials, rospy\_tutorials, and turtlesim packages from the ROS tutorials \cite{ros_tutorials} and the roslaunch package from the roscomm repository \cite{ros_comm}. A custom package containing launch files with known errors was also tested. Using the Import Launch plugin and then using the Export Launch plugin should make no functional changes to the file. Changing the whitespace, the order of the tag attributes, or the order of the sibling tags should have no effect on running the file, so these are considered acceptable changes. Table \ref{tab:import-export} shows a summary of the results of testing on these packages.

There were some changes after importing and exporting the files. The order of the sibling tags and the whitespace was changed. The order of attributes within a tag was also frequently changed. Some tags that were self-closing in the original file were translated into explicit open and closed tags. Any comments that were present in the original file were lost. These changes should have no functional effect. There was one large change noted that may affect the execution of the launch file. One of the original files had an ssh-port attribute included in the machine tag that was not preserved in the output launch file. This file came from the roslaunch package \cite{ros_comm}. In the official ROS wiki \cite{roslaunch_xml}, ssh port is not listed as a valid attribute for the machine tag, so it was not included in the tool. Since the attribute is not in the official documentation, this difference is acceptable. If the attribute was needed, it could be added manually to the exported file.

In the case of duplicate tags, the order of tags can have an effect on the execution of a launch file. For instance, if two arguments are declared with the same name but different values, the one that appears later in the launch file will take precedence. In this scenario, changing the order of the tags changes the result. The tool currently does not provide a way for users to choose the order of sibling tags. However, since including duplicate tags that overwrite each other's values is generally considered a bad practice, this edge case can be safely ignored.

The tool was also tested for importing and exporting launch files that contain known errors. The three error cases that were tested were a nonexistent tag, a nonexistent attribute, and a tag that is not closed. In the cases with a nonexistent tag and a tag that is not closed, attempting to import the file fails, and no launch file is created. The case with an attribute that does not exist did allow the attribute to be created. In this case, an exclamation mark is displayed next to the invalid attribute, and hovering over the message reveals a tooltip that says ``Remove META-invalid property.'' Clicking the symbol then removes the invalid attribute. If the user does not remove the invalid attribute and then exports the file, the invalid attribute will not be included in the exported launch file. If the desired behavior is to keep this attribute, the attribute would have to be added manually after exporting the file.

\begin{table}[htbp]
\caption{Summary of ROS Repositories}
\label{tab:import-export}
\begin{tabular}{@{}lll@{}}
\toprule
Package & Count of Launch Files & \makecell{Count of Functionally Identical\\ Launch Files After Import and Export} \\

\midrule
\texttt{roslaunch} \cite{ros_comm} & 14 & 13 \\
\texttt{roscpp\_tutorials} \cite{ros_tutorials} & 1 & 1 \\
\texttt{rospy\_tutorials} \cite{ros_tutorials} & 17 & 17 \\
\texttt{turtlesim} \cite{ros_tutorials} & 1 & 1 \\
\texttt{bad\_package} (custom package) & 3 & 0 \\
\bottomrule
\end{tabular}
\end{table}

\subsection{Runtime Analysis}
For the runtime analysis of each of the plugins, the notation shown in Table \ref{tab:runtime_notation} is used. When the subscript 0 is used, this indicates the initial quantity of an element in the model.

\begin{table}[htbp]
\caption{Notation for Plugin Runtime Analysis}
\label{tab:runtime_notation}
\begin{tabular}{@{}ll@{}}
\toprule
Symbol & Meaning \\
\midrule
$N$ & Number of all types of tags \\
$M$ & Number of \texttt{Node} tags \\
$A$ & Number of \texttt{Argument} tags \\
$T$ & Number of \texttt{Test} tags \\
$E$ & Number of edges in the argument dependency graph \\
$P$ & Total number of \texttt{Publisher} elements \\
$p$ & Average number of \texttt{Publisher} elements per \texttt{Node} \\
$S$ & Total number of \texttt{Subscriber} elements \\
$s$ & Average number of \texttt{Subscriber} elements per \texttt{Node} \\
$I$ & Number of \texttt{Include} tags \\
$C$ & Number of \texttt{Topic} connections \\
$P_{group}$ & Total number of \texttt{GroupPublisher} elements \\
$p_{group}$ & Average number of \texttt{GroupPublisher} elements per \texttt{Group} \\
$S_{group}$ & Total number of \texttt{GroupSubscriber} elements \\
$s_{group}$ & Average number of \texttt{GroupSubscriber} elements per \texttt{Group} \\
$G$ & Number of \texttt{Group} tags \\
$R$ & Number of \texttt{Remap} tags \\
$M_{library}$ & Number of \texttt{Node} tags in \texttt{Node Library} \\
$T_{library}$ & Number of \texttt{Test} tags in \texttt{Test Library} \\ 
$I_{library}$ & Number of \texttt{Include} tags in \texttt{Include Library} \\
\bottomrule
\end{tabular}
\end{table}

\subsubsection{Update Library}
The Update Library plugin initially deletes the initial \texttt{Node Library}, \texttt{Test Library}, and \texttt{Include Library} in $O(M_{library,0} + T_{library,0} + I_{library,0})$ time. Parsing the input configuration file to create new library elements with the appropriate publishers and subscribers takes $O(M_{library}(p+s) + T_{library}(p+s) + I_{library}(p_{group}+s_{group}))$ time. It should be noted that the number of tests and nodes added to the library will always be equal. The total runtime of the plugin is therefore $O(M_{library,0} + T_{library,0} + I_{library,0} + M_{library}(p+s) + T_{library}(p+s) + I_{library}(p_{group}+s_{group}))$.

\subsubsection{Import Launch}
The Import Launch plugin begins execution by building a dictionary of all library elements in time $O(M_{library} + T_{library} + I_{library})$. It then iterates through all tags in the imported launch file and creates corresponding model elements in $O(N)$ time. For the \texttt{Node}, \texttt{Test}, and \texttt{Include} tags, additional work may be required to copy over publishers and subscribers from the library. This is done in $O(M(p+s)+T(p+s)+I(p_{group}+s_{group}))$ time. The total work to run this plugin is therefore $O(M_{library} + T_{library} + I_{library} + N + M(p+s)+T(p+s)+I(p_{group}+s_{group}))$.

\subsubsection{Make Connections}
The Make Connections plugin begins by recording a list of all elements present in the model in $O(N)$ time. The plugin then deletes existing \texttt{Topic} connections, \texttt{GroupPublisher} elements inside of \texttt{Group} elements, and \texttt{Groupsubscriber} elements inside of \texttt{Group} elements in $O(C_0 + P_{group,0} + S_{group,0})$ time. The next step is to add the appropriate \texttt{GroupPublisher} and \texttt{GroupSubscriber} elements to the model. First, the \texttt{Group} elements are sorted in $O(G \lg G)$ time to traverse from the bottom to the top level tags. Within each \texttt{Group}, nested publisher and subscriber elements are located, and appropriate remaps are applied in order to create the new publishers and subscribers. This operation takes $O(G \cdot R \cdot (P + S + P_{group} + S_{group}))$ time in total as a conservative upper bound. In practice, only nested publishers and subscribers are considered for each group, not every publisher and subscriber present in the model. At this point, a dictionary of all ports in the model is built in $O(P + S + P_{group} + S_{group})$ time. The \texttt{Remap} elements are sorted and then applied to the dictionary in time $O(R + R \lg R)$. Then, the \texttt{Topic} connections are created between all publishers and subscribers in an upper bound time of $O(G(P+P_{group})(S+S_{group}))$. In each group, only publishers and subscribers that are descendants of the group will be considered, so this limit is higher than the real execution time. The simplified sum of all these runtimes is $O(N + C_0 + P_{group,0} + S_{group,0} + G \lg G + G \cdot R \cdot (P + S + P_{group} + S_{group}) + R \lg R + G(P+P_{group})(S+S_{group}))$ for the entire plugin execution.

\subsubsection{Error Checking}
When running the Error Checking plugin, the first step is to traverse and record all elements in the model in $O(N)$ time. The plugin then collects all \texttt{Node} and \texttt{Test} elements in $O(M+T)$ time and checks for duplicate names in time $O((M+T)^2)$. The \texttt{Argument} elements are each checked for correct attribute definitions in $O(A)$ time. The arguments are checked for circular dependencies in $O(A+E)$ time using a topological sort. The total runtime is then $O(N + M + T + (M+T)^2 + A + A + E)$. This simplifies to $O(N + (M+T)^2)$ since it is true that $A \le N$, and $T \le N$, and $M \le N$. Given that in most cases there are a small number of arguments and a small argument dependency graph, generally it can also be assumed that $E \le N$.

\subsubsection{Export Launch}
The Export Launch plugin processes every tag exactly once in a depth-first traversal order, taking $O(N)$ time. For the tags that have children, two sorts occur. All of the children are sorted in the general sort once. Also, just the argument tag children are sorted topologically to ensure that dependency ordering is maintained. Each tag will be sorted once in a general tag-priority sort. The sum of these sorts across all tags is bounded by $\sum _{\text{all tags}} n_i \lg n_i \le O(N \lg N)$, where $n_i$ is the number of children of the $i^{th}$ tag. The topological argument sort also only occurs once per tag, so the total work of these sorts is bounded by $O(A+E)$. Therefore, the work of all these sorts can be aggregated to $O(N \lg N + A + E)$. The total work of the plugin then is $O(N \lg N + A + E)$.

\section{Limitations} \label{section:limitations}

The tool does have some limitations. It does not have any support for visualizing services or actions. Services are synchronous communication between nodes in the form of a client-server relationship \cite{ros_services_wiki}. Actions are another form of communication between nodes that requires a goal, feedback, and result. They work well for defining long running tasks \cite{weon_millan_foxglove_ros1_actions_2022}. The only communication shown in ROSLaunchVisual is the relationship between publishers and subscribers, not services or actions. Visualizing these other forms of communication could be helpful in future work.

Another limitation of the tool is that it does not support ROS 2 launch files. ROS 2 launch files can be very different from ROS 1 launch files \cite{ros_migrating_launch_2025}. While many of the same tags are available in both versions, the attributes that belong to these many of these tags have changed. For instance, in the node tag, the ROS 1 ``type'' and ``ns'' attributes have become ``exec'' and ``namespace'' attributes. In some cases, attributes have been added or subtracted. The nesting rules have changed. For example, in ROS 1, parameters can be defined globally. In ROS 2, parameters can only be used if nested inside a node tag. There are also some ROS 1 tags that are not supported and some ROS 2 tags that have been newly created. For instance, at the moment, the machine and test tags from ROS 1 are not supported in ROS 2. Also, ROS 2 includes new tags, such as the let tag and the executable tag. With such large changes, it is highly unlikely that a launch file built with ROSLaunchVisual would function as intended in ROS 2. The metamodel and plugins are designed specifically with the rules for ROS 1 in mind. In addition to the XML changes, in ROS 2, launch files can also be Python files or YAML files. In the current state of the tool, only XML files can be imported and exported. Furthermore, the ROS Configuration Generator installs ROS 1 and assumes that all packages are ROS 1 packages. ROS 2 packages would fail to build and therefore could not be analyzed. It may be useful in the future to develop a new version of the tool to support ROS 2, but the current version can only support ROS 1.

Also, when a launch file has an include tag, the details of the included launch file are not translated into a diagram format. For instance, consider the following section of a launch file shown in Listing \ref{code:include}.

\FloatBarrier
\begin{samepage}
\lstset{
    language=xml,
    frame=single,
    framexleftmargin=5pt,
    framexrightmargin=5pt,
    framesep=12pt,
    tabsize=3,
    caption=include.launch,
    label=code:include,
    rulesepcolor=\color{gray},
    commentstyle=\color{OliveGreen},
    stringstyle=\color{red},
    numbers=left,
    numberstyle=\tiny,
    numbersep=5pt,
    breaklines=true,
    showstringspaces=false,
    basicstyle=\ttfamily\footnotesize,
    emph={},emphstyle={\color{magenta}},
}
\begin{lstlisting}
<launch>
    <include file="$(find pkg-name)/nested.launch">
        <arg name="a" value="value"/>
    </include>
</launch>
\end{lstlisting}
\end{samepage}
\FloatBarrier

Assume that nested.launch is the following file shown in Listing \ref{code:nested}. The listener node has one subscriber to the topic /msg. The talker node has one publisher to the topic /msg.

\FloatBarrier
\begin{samepage}
\lstset{
    language=xml,
    frame=single,
    framexleftmargin=5pt,
    framexrightmargin=5pt,
    framesep=12pt,
    tabsize=3,
    caption=nested.launch,
    label=code:nested,
    rulesepcolor=\color{gray},
    commentstyle=\color{OliveGreen},
    stringstyle=\color{red},
    numbers=left,
    numberstyle=\tiny,
    numbersep=5pt,
    breaklines=true,
    showstringspaces=false,
    basicstyle=\ttfamily\footnotesize,
    emph={},emphstyle={\color{magenta}},
}
\begin{lstlisting}
<launch>
    <node name="listener" pkg="my_pkg" type="listener.py"/>
    <node name="talker" pkg="my_pkg" type="talker.py"/>
    <arg name="b" value="value2"/>
    <remap from="/msg" to="/chatter"/>
</launch>
\end{lstlisting}
\end{samepage}
\FloatBarrier

In the visual representation of the file in Listing \ref{code:include}, inside the include tag, there would be three elements: a \texttt{GroupPublisher} with node name ``talker'' and topic ``/chatter'', a \texttt{GroupSubscriber} with node name ``listener'' and topic ``/chatter'', and an \texttt{Argument} with name ``a'' and value ``value''. The only information about the contents of nested.launch included in the diagram for include.launch is the publishers and subscribers started by the file. There is no information about argument b, the remap tag, or any other tags included in the file. With these publishers and subscribers from the included file, it is not possible to tell if these nodes were within a group, if they have been remapped, or if they were started as part of another included launch file. They are only observable as seen from a black-box analysis of the file provided by the ROS Configuration Generator. The program was designed this way so that users would not attempt to edit the contents of an included launch and instead only focus on how the static file could fit within the current context of the launch file being built with the diagram.

\section{Related Work} \label{section:relatedwork}

Modeling and visual programming languages are commonly used in the field of robotics to help simplify the design process. Abstracting the design to a higher level allows developers to focus on overall design goals rather than on smaller details like syntax. The following section describes other research in visual programming for robotics and other modeling tools that generate ROS launch files.

\subsection{Model-Based Generation of ROS Launch Files}

There are several tools available that apply modeling to the development of ROS launch files. Table \ref{tab:visual_comparison} summarizes the features of several of these tools compared to ROSLaunchVisual. Most of the tools are designed to generate launch files in ROS 1, while one tool is designed for ROS 2. The majority of the tools provide a graphic representation of the model for editing, while one uses a text-based model. Two of the tools can import an existing launch file for editing. Only two of the tools discuss automatically parsing the ROS code to identify publishers and subscribers and using this information in the modeling process. Some of the tools provide a simulated example of running the finished file, unlike ROSLaunchVisual. Although most of the tools can be used to build any generic launch file, two are designed to build launch files for more specific use cases.

\begin{sidewaystable}[htbp]
\caption{Comparison of Launch File Modeling Tools}
\label{tab:visual_comparison}
\begin{tabular}{@{}lccccccc@{}}
\toprule
Tool & ROS 1 / ROS 2 & Visual Model & Import & Dynamic ROS Parsing & Simulation & General-Purpose \\
\midrule
rxDeveloper \cite{mullers2012rxdeveloper}  & ROS 1 & Yes & -  & -  & Yes & Yes \\
RADOE \cite{narayanamoorthy2015creating} & ROS 1 & Yes & Yes  & -  & Yes & Yes \\
ROSSi \cite{9649736} & ROS 2 & Yes & - & Yes  & Yes & Yes \\
ReApp \cite{wenger2016model} & ROS 1 & Yes & -  & - & - & -\\
Hua et al. \cite{7733579} & ROS 1 & - & - & - & - & -\\
ROSLaunchVisual & ROS 1 & Yes & Yes & Yes & - & Yes \\
\bottomrule
\end{tabular}
\end{sidewaystable}

The tool rxDeveloper \cite{mullers2012rxdeveloper} uses a GUI to aid in ROS software development. Launch files and parameter files can both be generated using this tool. Developers can build computational graphs with node communication that are then easily translated into code from the model. Many XML launch tags are supported. rxDeveloper also allows users to provide a node specification file that contains details about publishers, subscribers, parameters, and services. However, there is no method for automatically generating this node specification file, so someone with knowledge of the system must manually create it. Unlike ROSLaunchVisual, rxDevloper cannot import existing launch files or dynamically parse ROS files to extract topic information.

Within RADOE (Robot Application Development and Operating Environment) \cite{narayanamoorthy2015creating}, there is a visual programming tool that edits ROS launch files. In the tool, different ROS launch XML tags are added to the visual model. Arrows can be toggled on and off to demonstrate the potential communication between different nodes. The tool can pull information about node publishers and subscribers from a provided XML file. The developer also has the option of adding publishers and subscribers as nodes are added to the model. However, this tool cannot generate this publisher and subscriber information automatically like the ROS Configuration Generator does. Existing launch files can be loaded into the model so that they can be edited more easily in a visual format. The tool also provides an option to run the file temporarily on a terminal emulator for testing purposes. The initial use case study for this tool revealed that using a visual programming tool was preferred over editing a textual XML launch file.

ROSSi (ROS - Simple) \cite{9649736} is designed to generate ROS 2 code and launch files. One use case of the tool is generating skeleton code for ROS 2 nodes in the form of a setup phase, loop phase, and destroy phase. The tool is also useful for generating Python launch files to run in ROS 2, a feature that is not currently included in ROSLaunchVisual. Only ROS 1 functionality is included in launch files created with ROSSi, such as nodes, included launch files, and namespaces. The authors mention the possibility of using black-box testing to automatically discover the publishers and subscribers in a node, while also highlighting the limitations of this method. However, no details are provided on the implementation of this method within ROSSi. ROSSi also provides the ability to view a live diagram of the execution of the system specified by the launch file. Since ROSLaunchVisual is focused mainly on designing launch files rather than testing them live, this feature is not included in ROSLaunchVisual.

ReApp (Reusable Robot Applications for Flexible Robot Plants Based on Industrial ROS) \cite{wenger2016model} is a workbench that allows the design of model-based robots for applications with reusable components. It is specifically designed to help program industrial robots for a variety of different vendors and use cases. The Skill and Solution Modeling Tool allows users unfamiliar with ROS to compose a ROS system graphically and export the result to a launch file. The editor displays nodes as boxes and allows designers to draw logical connections between them. Unlike ROSLaunchVisual and some of the other tools presented in this section, ReApp cannot be used to generate any arbitrary launch file. It is specifically designed to generate files for the industrial model use case. By using this robot application model and a graphical interface, it is easier for novice developers to build new applications for industrial robots by reusing existing code.

The tool presented in \cite{7733579} is another example of a tool that generates ROS launch files from a modeling approach for a specific use case. This approach uses AutomationML as a modeling framework to build ROS robotic applications. Rather than using a visual model, this tool builds the model in a text format and then translates it to an XML launch file. The specific use case in this paper is industrial robots. Many of the other tools, like ROSLaunchVisual, instead build launch files for generic use cases. Although the language presented in \cite{7733579} is not useful for building a launch file for every use case, it simplifies the process for a specific domain.

\subsection{Other Related Work}

The BRICS Component Model \cite{bruyninckx2013brics} is a project that assists robotics developers with guidelines, metamodels, tools, and other structures. The metamodels proposed in this paper are platform-independent and do not go into any application-specific details. Rather than code generation, the goal of this modeling method is to provide structure in the engineering development process and the design of robotic systems.

ROSMOD (Robot Operating System Model-driven development tool suite) \cite{electronics5030053} is another tool that uses WebGME to generate code for ROS. This tool is used to rapidly develop and test ROS code. The authors present a case study where ROSMOD was used to develop code for an Autonomous Ground Support Equipment (AGSE) robot for the 2014-2015 NASA Student Launch Competition. Using this modeling solution, the authors were able to produce successful ROS code in a short amount of time with a limited number of team members.

In \cite{karaca2020ros}, a visual programming language is proposed to generate ROS code for EvaRobot. The web based visual programming system allows developers to chain blocks together to write code that creates publishers and subscribers and defines more high-level ROS behaviors like teleoperation or wandering. Rather than building launch files, this programming language allows users to program robots at a more abstract level without having to worry about syntax errors. 

In \cite{10.1007/978-3-642-34327-8_14}, the authors use AToM\(^3\) (A Tool for Multi-formalism Meta-Modeling) to develop a domain-specific modeling language that generates ROS code. Specifically, the model generates skeletons of code to define ROS nodes and their publishers and subscribers. The logic that lives within the nodes that creates messages for the publishers and subscribers cannot be generated. This tool provides another example of simplifying code development for robotic systems with visual modeling.

Another visual programming language for ROS is described in \cite{Vázquez_Calvo_Fernández_Ramos_2021}. A block-based language is presented that can write code for many different types of robots, including a manipulator, humanoid, snake, hexapod, and rover robot. Each action that the robot should perform is represented by a block, making it easy for beginners to program a robot. The ROS controller code is automatically generated, letting developers focus more on high-level design.

All of these projects provide helpful contributions in areas of robotics and metamodeling.  The BRICS Component Model provides helpful methods of abstraction without implementation. The other works focus on generating code for node components that fit within the overall system. The focus of ROSLaunchVisual, instead, is creating models for the system-level configurations and generating configuration files.

\section{Conclusion} \label{section:conclusion}

ROSLaunchVisual is a useful tool for the model-based development of ROS launch files. Writing these files in a textual format can be hard to understand and prone to syntax errors. The tool provides a visual representation of the hierarchy and relationship between elements of the file. The plugins aid in the development process and speed up the overall design time through importing existing files, checking for errors, and exporting finished XML files that can be executed in the robotic system.

One of the most crucial contributions of this work is the ability to see how nodes communicate with each other, as this is not possible with the textual representation of a launch file. The ROS Configuration Generator provides a dynamic analysis of ROS code to extract information about publishers and subscribers. A modeling approach to launch files provides a simple way for users to import and create launch files, and then rapidly design and deploy these files into real use cases.

\bmhead{Acknowledgements}

This material is based upon work supported by the National Science Foundation under Grant No. 2135579. Additional thanks are given to Tamas Kecskes and Janos Sztipanovits for their instructive feedback on topics that laid the foundations for this work.

\bibliography{sample-base}
\bibliographystyle{sn-mathphys-num}

\end{document}